\documentclass{article}

\usepackage[final]{hydrahead}
\usepackage{multirow}

\usepackage[utf8]{inputenc} 
\usepackage[T1]{fontenc}    
\usepackage{hyperref}       
\usepackage{url}            
\usepackage{booktabs}       
\usepackage{amsfonts}       
\usepackage{nicefrac}       
\usepackage{microtype}      
\usepackage{xcolor}         
\usepackage{amsmath}
\usepackage{graphicx}
\usepackage{xeCJK}
\usepackage{float}
\usepackage{subcaption}
\usepackage{tablefootnote}
\usepackage{eso-pic}
\usepackage{graphicx}

\usepackage[most]{tcolorbox}

\definecolor{abstractbg}{HTML}{F3F0FF}    
\definecolor{abstractborder}{HTML}{C4B5FD} 
\definecolor{abstracttitle}{HTML}{5B21B6}  

\newtcolorbox{techreportabstract}{
    colback=abstractbg,           
    colframe=abstractborder,      
    arc=6pt,                      
    boxrule=1pt,                  
    left=18pt, right=18pt,       
    top=15pt, bottom=15pt,        
    halign=justify,               
    before=\vspace{10pt},         
    after=\vspace{15pt}          
}

\usepackage{fancyhdr}

\usepackage[style=iso]{datetime2} 

\fancypagestyle{firstpage}{
    \fancyhf{} 
    \fancyhead[R]{\textcolor{gray}{\sffamily\raisebox{-25pt}{\today}}} 
     
}

\AddToShipoutPictureBG*{%
  \AtPageUpperLeft{%
    \hspace{2.5cm}%
    \raisebox{-2.5cm}{%
      \includegraphics[width=1.0cm]{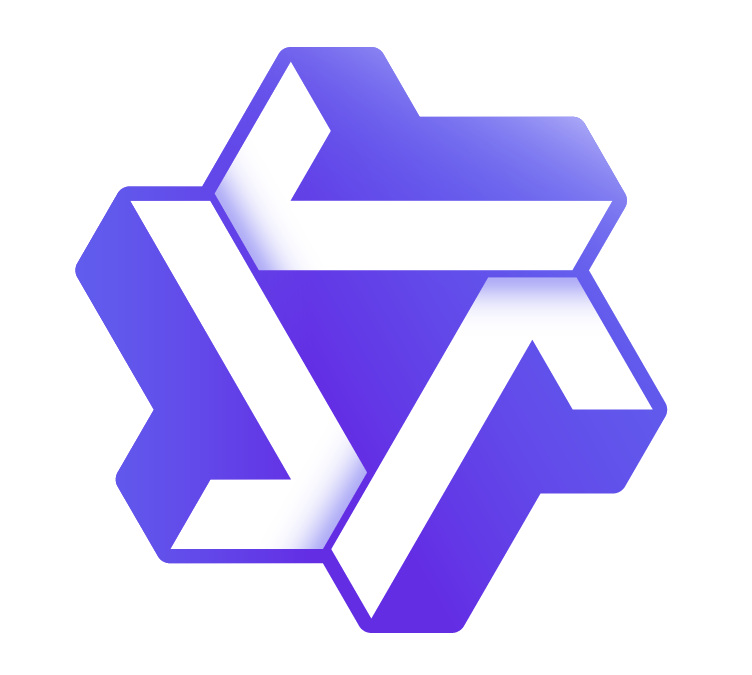}%
    }%
  }%
}

\title{HydraHead: From Head-Level Functional Heterogeneity to Specialized Attention Hybridization}

\author{
Zhentao Tan, Wei Chen, Jingyi Shen, Yao Liu, Xu Shen, Yue Wu, Jieping Ye \\
Alibaba Group
}

\usepackage[table]{xcolor}

\begin{document}

\maketitle

\thispagestyle{firstpage} 

\begin{techreportabstract}
    {\centering\large\bfseries Abstract\par}
    \vspace{0.8em} 

The quadratic complexity of attention poses a critical bottleneck for long-context processing, spurring interest in hybrid attention designs. Most open-source hybrid models adopt a layer-wise strategy. Yet, prior work has noted the inherent difficulty of integrating Linear Attention (LA) with Full Attention (FA), suggesting that the design space of attention hybridization remains underexplored.
To probe this space, we conduct interpretability analysis and observe that layers exhibit block-wise functional similarity, while individual heads within the same layer display distinct functional specialization despite sharing input features. This head-level heterogeneity suggests that the head dimension provides a natural and principled granularity for fusing heterogeneous attention signals.
Building on this insight, we introduce \textbf{HydraHead}, a novel architecture that hybridizes FA and LA along the head axis. HydraHead features two key innovations: (1) an interpretability-driven selection strategy that identifies retrieval-critical heads and preserves FA only for them, and (2) a scale-normalized fusion module that reconciles the distributional gap between FA and LA head outputs. By leveraging a three-stage transfer pipeline with parameter reuse and distillation, we achieve high-performance hybrid models with minimal training overhead.
Under a unified training setup, HydraHead outperforms other hybrid designs in long-context tasks while maintaining strong general reasoning.
With interpretability-driven head selection, it matches a 3:1 layer-wise hybrid's long-context performance at a 7:1 LA-to-FA ratio. Crucially, trained on only 15B tokens, HydraHead achieves over \textbf{69\%} improvement over the baseline at 512K context length, approaching Qwen3.5, a leading model of comparable size with a native context length of 256K. This highlights the significant scaling potential of head-level hybridization.
\end{techreportabstract}

\begin{figure}[H]
    \centering
    \includegraphics[width=1\linewidth]{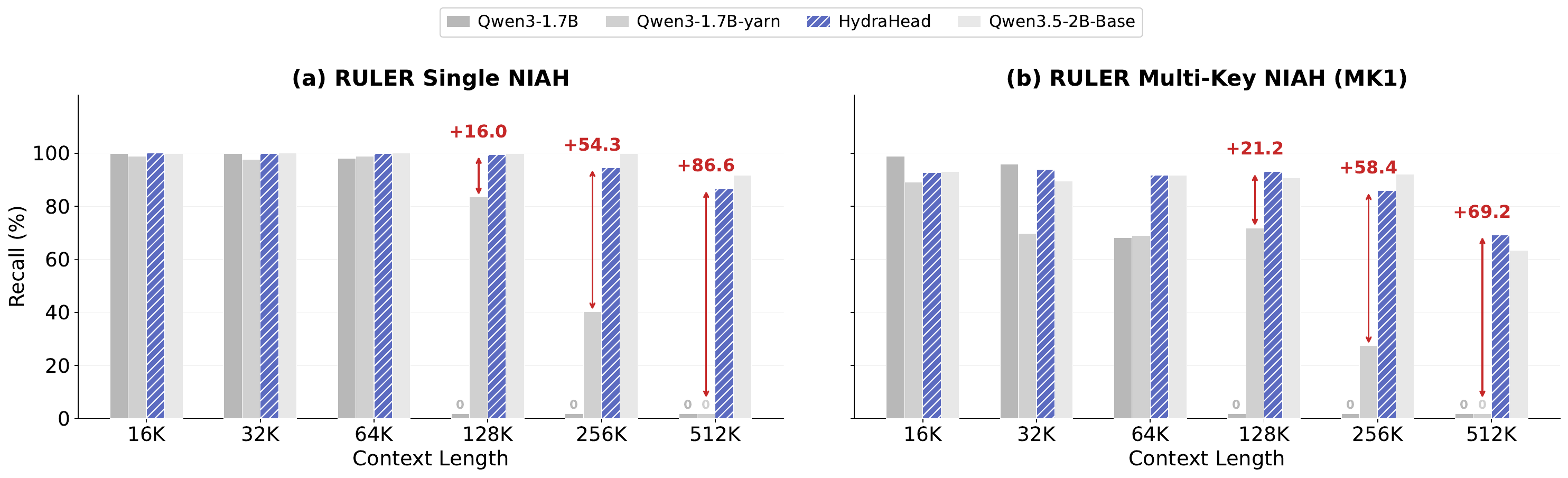}
    \caption{HydraHead, trained on only 15B tokens, significantly outperforms the Qwen3-1.7B baseline on NIAH tasks, approaching Qwen3.5-2B-Base, a leading model of comparable scale with native 256K support.}%
    \label{fig:teaser}
\end{figure}

\section{Introduction}

The landscape of Large Language Models (LLMs) has witnessed a pivotal transition from static question-answering systems to autonomous agents capable of complex reasoning and long-horizon planning~\cite{glm5team2026glm5vibecodingagentic,kimiteam2026kimik2openagentic,qwen2026qwen35blog,deepseek2026v4}. This evolution places unprecedented demands on context window extension, yet the quadratic complexity of standard Full Attention (FA) remains a rigid computational barrier. While Linear Attention (LA) paradigms—often rooted in State Space Models (SSMs)~\cite{mamba,mamba2,lahoti2026mamba3improvedsequencemodeling} or kernel-based approximations~\cite{schlag2021lineartransformerssecretlyfast,yang2025gated,kimiteam2025kimilinearexpressiveefficient,hatamizadeh2026gateddeltanet2decouplingerase}—offer linear-time complexity, they frequently suffer from ``expressivity collapse," struggling to maintain the high-precision retrieval required for sophisticated reasoning tasks.

Driven by this tension, many recent LLMs adopt hybrid architectures—predominantly layer-wise designs that interleave different attention mechanisms across layers—integrating linear with softmax attention~\cite{qwen2026qwen35blog,kimiteam2025kimilinearexpressiveefficient,minimax2025minimaxm1scalingtesttimecompute}, or blending full with sliding-window attention~\cite{coreteam2026mimov2flashtechnicalreport}. These efforts reflect a shared direction: scaling long-context capabilities is no longer a matter of selecting a single attention mechanism, but an exercise in architectural fusion. Despite these successes, some prior work has noted that training hybrid models that combine linear attention with full attention remains challenging, with efforts ultimately focusing on full- or sparse-attention variants~\cite{glm5team2026glm5vibecodingagentic,minimax2026minimaxm2seriesminiactivations,deepseek2026v4}, suggesting that the design space of attention hybridization remains underexplored.

\begin{figure}[H]
    \centering
    \begin{subfigure}[b]{0.61\linewidth}
        \centering
        \includegraphics[width=\linewidth]{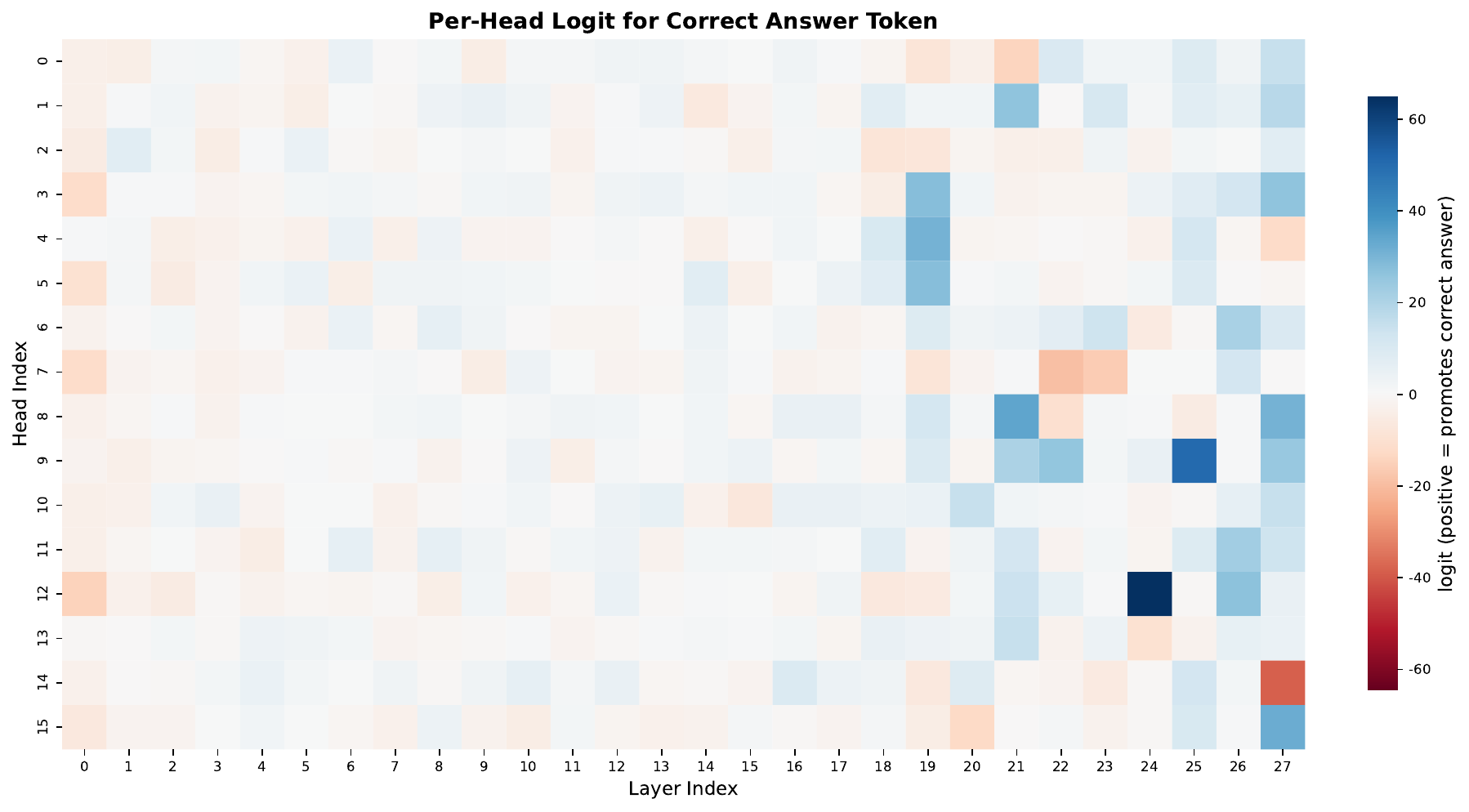}
        \caption{Per-head logit for correct answer token.}%
        \label{fig:comm_headlogit}
    \end{subfigure}
    \hfill
    \begin{subfigure}[b]{0.37\linewidth}
        \centering
        \includegraphics[width=\linewidth]{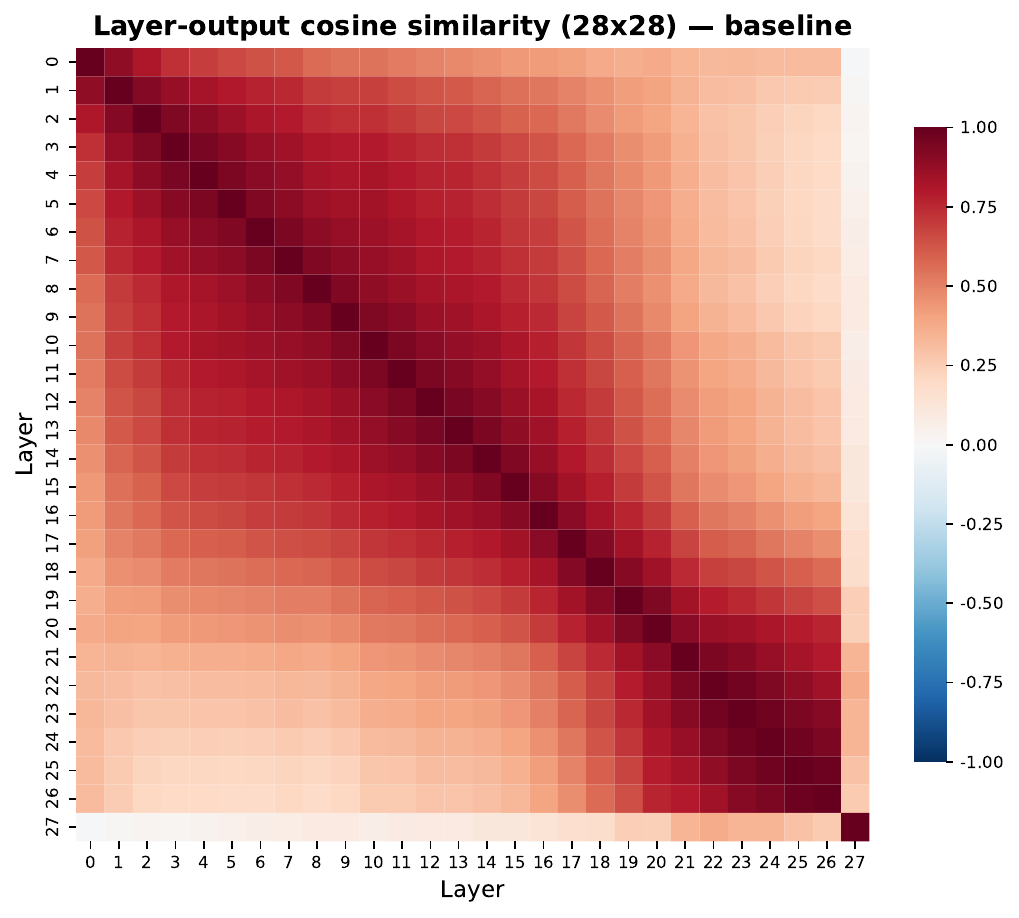}
        \caption{Layer-output cosine similarity.}%
        \label{fig:comm_layersim}
    \end{subfigure}
    \caption{\textbf{The head is a natural granularity for hybridization.} (a) Individual heads within the same layer are \emph{functionally heterogeneous}, with per-head contributions to the final output varying substantially. (b) Layer-level outputs vary smoothly across depth, making it difficult to identify clear boundaries for assigning distinct attention mechanisms to different layers.}%
    \label{fig:comm_visual}
\end{figure}

To navigate this underexplored design space, it is essential to look beyond surface-level performance and examine how computation is internally organized. Mechanistic interpretability provides the necessary lens, having matured from small models to frontier-scale systems. Specifically, it now spans circuit-level analyses in small transformers~\cite{elhage2021mathematical,wang2022interpretabilitywildcircuitindirect,gao2025weightsparse}, scalable feature extraction in advanced models from Anthropic~\cite{templeton2026scaling}, OpenAI~\cite{gao2025scalingsae}, and Google DeepMind~\cite{lieberum2024gemmascope}, as well as comprehensive anatomical studies inside frontier LLMs~\cite{lindsey2025biology,lindsey2026introspective,sofroniew2026emotion}. Importantly, a recurring finding across these studies is that attention heads exhibit substantial diversity in both functional role and importance~\cite{voita2019analyzingmultiheadselfattentionspecialized,michel2019sixteenheadsreallybetter,clark2019doesbertlookat}, and this understanding has been extended to diagnose and modify targeted model behaviors~\cite{zhang2024interpretingimprovinglargelanguage,chen2025yesmentruthtellersaddressingsycophancy}, providing an analytical foundation for grounding architectural design in the internal structure of pretrained models.

We examine two candidate granularities for attention hybridization: the attention head, one of the most commonly studied functional units in interpretability research, and the layer, which mainstream hybrids predominantly adopt. At the head level, the analysis reveals sharp functional heterogeneity. Figure~\ref{fig:comm_headlogit} visualizes per-head logit contributions~\cite{elhage2021mathematical} to the correct answer token on a long-context retrieval task: within a given layer, only a sparse subset of heads contributes substantially while many heads in the same layer remain nearly inactive. These retrieval-critical heads concentrate in deeper layers, consistent with the head-level functional diversity noted above and with prior findings that deeper layers carry more task-specific signals~\cite{wang2022interpretabilitywildcircuitindirect,DBLP:conf/emnlp/GevaBFG23}. These observations are consistent with the head-level diversity noted above. At the layer level, by contrast, the output similarity matrix exhibits a smooth block-wise structure (Figure~\ref{fig:comm_layersim}), where similarity decays gradually across adjacent layers---a pattern consistent with earlier findings~\cite{kornblith2019similarityneuralnetworkrepresentations,raghu2022visiontransformerslikeconvolutional}. This smooth variation does not preclude layer-wise hybridization, but offers limited discriminative signal for deciding where to place different attention mechanisms. Together, these observations suggest a natural design principle: the head, rather than the layer, provides a functionally grounded and sufficiently fine-grained unit for attention hybridization, enabling different heads within the same layer to process identical input through distinct mechanisms in a shared representation space.

Building on these insights, we propose \textbf{HydraHead}, a fine-grained head-level attention hybridization architecture that addresses two core challenges: (1) determining which heads should adopt FA versus LA, and (2) effectively fusing the heterogeneous attention patterns produced by different mechanisms. For the first challenge, we leverage interpretability techniques to identify heads critical for precise token-level retrieval directly from a pretrained model, retaining FA exclusively for these heads while assigning LA to the remainder for efficient long-context scaling. This head selection procedure is lightweight and one-shot, requiring only a few forward passes over a small calibration set. For the second challenge, we mix FA and LA at the head dimension within each layer, assigning each attention head an independently configurable attention mechanism. Owing to the paradigm gap between FA and LA in how attention features are formed~\cite{meng2026stillselectingtokensintralayer,meng2026normtimesdirectionrestoringmissingquery}, we introduce a head-wise scale-normalized fusion to mitigate interference when these heterogeneous attention signals are mixed. To train the resulting hybrid model, we introduce a three-stage transfer pipeline: parameter inheritance from the pretrained checkpoint and layer-wise alignment; global distillation; and long-context fine-tuning.

We evaluate HydraHead against several representative hybridization strategies, all built upon Qwen3-1.7B~\cite{yang2025qwen3technicalreport} under a unified training configuration, using both long-context retrieval benchmarks at varying sequence lengths, and general reasoning benchmarks spanning multiple difficulty levels. These baselines span three dominant paradigms: layer-wise hybrids, token-wise hybrids, and head-wise hybrids. Our experiments yield the following findings. 
\begin{itemize}
    \item \textbf{HydraHead achieves state-of-the-art long-context performance while maintaining advanced general reasoning capabilities.} Existing hybridization strategies reveal a consistent trade-off across granularity levels: compared to layer-wise hybrids, both head-wise and token-wise approaches excel at general reasoning but struggle with length extrapolation. As a fine-grained head-wise approach, HydraHead retains a robust \textbf{>10\%} gain over layer-wise baselines on hard reasoning tasks, while achieving \textbf{state-of-the-art} long-context performance. 
    \item \textbf{Interpretability-based head selection enables aggressive FA compression with minimal performance degradation.} By effectively identifying retrieval-critical heads, our method matches the overall performance of a 3:1 layer-wise hybrid even at substantially higher LA-to-FA mixing ratios (e.g., 7:1).
\end{itemize}

Beyond controlled comparisons, we further challenge HydraHead by directly benchmarking against state-of-the-art models. As shown in Figure~\ref{fig:teaser}, through meticulous optimization of training hyperparameters and data composition, our model achieves substantial long-context capabilities with only 15B training tokens. Specifically, it delivers over \textbf{69\%} improvement on NIAH benchmarks at 512k context compared to the pretrained baseline, approaching the performance of Qwen3.5~\cite{qwen2026qwen35blog}, a leading model of comparable scale. Crucially, these gains are achieved without compromising general reasoning performance, further validating the significant potential of our head-wise hybrid method in efficient long-context modeling.

\section{Related Work}

\subsection{Mechanistic Interpretability and Head-Level Analysis}
Mechanistic interpretability aims to reverse-engineer the internal computations of neural networks through causal and structural analysis. Foundational work establishes a mathematical framework for reasoning about transformer circuits~\cite{elhage2021mathematical} and standardizes activation patching as the primary tool for measuring component-level causal effects~\cite{heimersheim2024useinterpretactivationpatching}. These techniques have enabled the identification of concrete circuits underlying specific behaviors, such as indirect object identification~\cite{wang2022interpretabilitywildcircuitindirect} and factual knowledge storage~\cite{meng2023locatingeditingfactualassociations}.

A recurring finding across this line of work is that attention heads serve as meaningful functional units. Voita et al.~\cite{voita2019analyzingmultiheadselfattentionspecialized} identify three specialized head types---positional, syntactic, and rare-word heads---and show that the majority of remaining heads can be pruned with negligible quality loss. Michel et al.~\cite{michel2019sixteenheadsreallybetter} confirm this finding broadly, demonstrating that in many tasks, only a small fraction of heads accounts for most of the model's capability. Clark et al.~\cite{clark2019doesbertlookat} further show that individual heads learn to attend to specific syntactic relations, revealing consistent functional specialization. At a deeper level, Wang et al.~\cite{wang2022interpretabilitywildcircuitindirect} decompose the IOI circuit in GPT-2 small into 26 heads organized into seven functional classes---the canonical worked example of head-level circuit discovery---establishing that head-level granularity is sufficient to capture complex multi-step computations.

Collectively, these findings establish attention heads as meaningful and well-studied functional units, and the associated causal tools---particularly activation patching and path patching---provide a principled methodology for quantifying their individual contributions. This body of work directly motivates and supports our investigation: we leverage these tools to analyze the functional structure of a pretrained model and use the resulting insights to guide the design of our hybrid attention architecture.

\subsection{Linear Attention}

Linear attention addresses the quadratic complexity of softmax attention by replacing the explicit token-to-token similarity computation with a recurrent state update, reducing complexity from $\mathcal{O}(T^2)$ to $\mathcal{O}(T)$. One prominent line of work frames this through the lens of state space models, progressively introducing input-dependent selectivity to overcome the rigidity of fixed dynamics~\cite{mamba,mamba2}, and more recently extending to complex-valued state spaces with higher-order discretization for richer memory representation~\cite{lahoti2026mamba3improvedsequencemodeling}. A complementary perspective views linear attention through the lens of fast-weight memory~\cite{katharopoulos2020transformersrnnsfastautoregressive}, where the key-value state is treated as an associative memory updated via the delta rule for precise credit assignment~\cite{schlag2021lineartransformerssecretlyfast}. Subsequent advances integrate gating mechanisms to control memory retention~\cite{yang2025gated}, enhance the expressiveness of the gating function itself~\cite{kimiteam2025kimilinearexpressiveefficient}, and most recently decouple the update and forget operations for finer-grained memory control~\cite{hatamizadeh2026gateddeltanet2decouplingerase}. Despite these advances, pure linear attention remains constrained by its fixed-dimensional recurrent state, which limits its capacity for long-range retrieval and exact token-level replication---motivating the hybrid designs that combine linear attention with full attention in a complementary fashion.

\subsection{Hybrid Transformers}

The pursuit of balancing computational efficiency and representational expressivity has catalyzed a diverse range of hybrid Transformer architectures. Existing efforts can be categorized into three primary paradigms based on the granularity of their hybridization.

\paragraph{Layer-wise Hybridization.} 
This paradigm alternates between different attention kernels across Transformer layers. Many works primarily focus on interweaving full-attention layers with efficient alternatives, such as windowed attention or linear approximations, to mitigate the quadratic memory footprint of long contexts~\cite{Fu2025NemotronFlashTL,nvidia2026nemotron3superopen,merrill2026olmohybridtheorypractice}. For instance, models such as MiMo~\cite{coreteam2026mimov2flashtechnicalreport}, MiniMax-M1~\cite{minimax2025minimaxm1scalingtesttimecompute}, and Qwen~\cite{qwen2026qwen35blog} systematically balance standard attention with efficient attention at configurable layer-wise ratios. Furthermore, automated strategies like those explored in GLM-5~\cite{glm5team2026glm5vibecodingagentic} utilize Neural Architecture Search (NAS) to discover optimal layer-wise configurations, demonstrating that static interleaving is often suboptimal. Despite the prevalence of layer-wise hybridization in state-of-the-art architectures, prior work has noted that training hybrid models combining linear attention with full attention remains challenging, with many efforts ultimately focusing on full or sparse attention variants~\cite{glm5team2026glm5vibecodingagentic,minimax2026minimaxm2seriesminiactivations,deepseek2026v4}, suggesting that the design space of attention hybridization remains underexplored.

\paragraph{Token-wise Hybridization.}
To achieve finer granularity, recent studies have proposed blending that operates at the token or sequence level~\cite{meng2026stillselectingtokensintralayer,du2026nativehybridattentionefficient,zhao2026switchattentiondynamicfinegrained,lan2025liger}. Among them, STILL~\cite{meng2026stillselectingtokensintralayer} employs saliency-based gating to decide which tokens warrant the precision of FA versus the efficiency of linear or sparse approximations. While these approaches refine the allocation granularity from the layer level to the token level, they face a potential challenge: maintaining stable and consistent sequence representations across long contexts is non-trivial when different tokens are processed through heterogeneous attention pathways. Few existing works have reported evaluation results beyond 32K context length for such methods. Moreover, the very pursuit of such stability and consistency may inadvertently drive different attention patterns toward convergence, thereby diminishing their ability to complement each other. Additionally, in the case of sparse attention, the set of tokens retrieved varies across queries, necessitating the retention of full KV caches and thus undermining memory efficiency during inference.

\paragraph{Head-wise Hybridization.} 
Recognizing the functional heterogeneity of multi-head attention, recent work has explored head-wise hybridization along two distinct paradigms. The first paradigm performs per-head selection: methods such as DuoAttention~\cite{xiao2024duoattentionefficientlongcontextllm} and Elastic Attention~\cite{tang2026elasticattentiontesttimeadaptive} treat each head as the unit of allocation, using heuristic partitioning or learned routers to designate ``critical'' heads for FA while assigning the remainder to sparse attention variants. However, since sparse attention and standard FA share the same quadratic dot-product computation core, their complementary benefits are inherently limited—both remain constrained by similar attention patterns, yielding marginal gains rather than a fundamental efficiency breakthrough. The second paradigm performs per-head mixing. For example, Hymba~\cite{dong2024hymbahybridheadarchitecturesmall} processes the input through two parallel branches—a full multi-head FA branch and a Mamba SSM branch—and fuses their output features via element-wise averaging.
Although these fusion-based approaches enrich representational capacity, they apply a uniform mixing strategy across all heads without accounting for their inherent functional specialization.
In contrast, our architecture leverages interpretability analysis to uncover the inherent functional specialization of different attention heads, and strategically assigns FA and LA to heads based on their complementary strengths. By promoting representational complementarity, it harmonizes the high-precision recall of FA with the constant-cost global context modeling of LA, achieving a synergistic fusion that fully exploits the distinct capabilities of each attention mechanism.

\subsection{Taming Transformer to Hybrid Architecture}

The prohibitive cost of pre-training hybrid architectures from scratch catalyzes a surge in cross-architecture distillation methods, aiming to transfer the reasoning capabilities of pre-trained Transformers into more efficient counterparts. Early attempts focus on distilling Transformers into pure RNN or State-Space Models (SSMs). For example, T2R~\cite{kasai2021finetuningpretrainedtransformersrnns} and RADLADS~\citep{goldstein2026radladsrapidattentiondistillation} demonstrate the feasibility of architectural conversion. However, MOHAWK~\citep{bick2025transformersssmsdistillingquadratic} and LoLCATs~\citep{zhang2025lolcatslowranklinearizinglarge} later highlight that naive distillation often leads to performance degradation on recall-intensive tasks, as the disparity between quadratic attention and linear-time mixers creates a significant knowledge transfer barrier.
To bridge this gap, recent works shift toward multi-stage pipelines and principled initialization strategies. HedgeMamba~\citep{moudgil2026attentionmambarecipecrossarchitecture} proposes a two-stage recipe that first distills softmax attention into a linearized kernel (Hedgehog) and then uses it to initialize an SSM-based Mamba mixer. Complementary approaches, such as KL-Guided~\citep{li2025distillinghybridattentionmodels}, emphasize the criticality of selective layer conversion to maximize long-context retention. We leverage the HALO~\citep{chen2026hybridlinearattentionright} as our foundation, adapting its efficient distillation procedure to support our novel head-wise hybrid structures. By extending HALO to handle head-level feature distribution shifts through decoupled $QKV$ projections and modulated fusion, our method successfully transitions from standard layer-wise conversion to a more granular, head-specific hybridization strategy.

\section{Preliminaries}
\label{sec:preliminaries}

\subsection{Causal Patching Methods}
\label{sec:ap_prelim}

\paragraph{Activation Patching.}
Activation patching~\cite{heimersheim2024useinterpretactivationpatching} is a causal intervention technique for quantifying the contribution of individual model components to a target behavior. The procedure requires a \emph{paired input}: a \emph{clean} input $x$ on which the model produces the correct behavior, and a \emph{corrupted} input $x'$ obtained by a minimal edit that alters the target answer. For a component of interest---e.g., the output $\mathbf{O}_{l,h}$ of attention head $h$ at layer $l$---the clean-run activation is replaced with its corrupted-run counterpart while all other components remain at their clean values. The resulting change in model behavior, measured by a scalar readout such as the \emph{logit difference}
\begin{equation}
    m(x) = z[a^{+}] - z[a^{-}],
    \label{eq:logit_diff_basic}
\end{equation}
where $z$ denotes the model's output logits, and $a^{+}$, $a^{-}$ are the correct and counterfactual answer tokens respectively, quantifies the component's causal necessity: a large drop indicates a causally indispensable component, while a negligible change indicates a dispensable one~\cite{wang2022interpretabilitywildcircuitindirect}.

\paragraph{Path Patching.}
While activation patching measures each component's \emph{direct} effect on the output, \emph{path patching}~\cite{wang2022interpretabilitywildcircuitindirect} enables finer-grained attribution by restricting the intervention to a specific computational path. Concretely, to measure how much an upstream head contributes \emph{through} a particular downstream head, one first runs the corrupted input and records the activations the upstream head sends to the downstream head; then runs an otherwise-clean forward pass but substitutes only those recorded activations at the downstream head's input. The resulting behavioral change isolates the upstream head's contribution along that specific path, revealing not only which components are important but also \emph{how} they contribute to each other.

\subsection{Grouped-Query Attention Mechanism}
\label{subsec:gqa}
Modern large language models predominantly adopt Grouped-Query Attention (GQA)~\cite{ainslie2023gqatraininggeneralizedmultiquery} to balance inference efficiency and modeling capacity. Thus, we consider a Transformer model with $L$ layers, where each layer consists of a GQA and a Multi-Layer Perceptron (MLP). Let $H$ denote the total number of query heads and $G$ denote the number of key-value heads, with $G < H$. The set of query heads $\mathcal{H} = \{1, \dots, H\}$ is partitioned into $G$ disjoint groups $\mathcal{G} = \{1, \dots, G\}$, such that all query heads in a group share a single key-value head.

Given an input sequence represented as a matrix $\mathbf{X} \in \mathbb{R}^{T \times d}$, where $T$ is the sequence length and $d$ is the hidden dimension, we define the head dimension as $d_h = d/H$. In standard GQA implementations, the key and value heads share the same dimension $d_h$ as the query heads. 

The attention mechanism employs three global learnable projection matrices: $\mathbf{W}^Q \in \mathbb{R}^{d \times d}$, $\mathbf{W}^K \in \mathbb{R}^{d \times d_G}$, and $\mathbf{W}^V \in \mathbb{R}^{d \times d_G}$, where $d_G = d \cdot \frac{G}{H}$ is the total dimension for all key/value heads. 
The global projected representations are computed as:
\begin{equation}
\mathbf{Q} = \mathbf{X} \mathbf{W}^Q \in \mathbb{R}^{T \times d}, \quad 
\mathbf{K} = \mathbf{X} \mathbf{W}^K \in \mathbb{R}^{T \times d_G}, \quad 
\mathbf{V} = \mathbf{X} \mathbf{W}^V \in \mathbb{R}^{T \times d_G}.
\end{equation}

These tensors are split along the feature dimension to obtain individual head representations. For each query head $h \in \mathcal{H}$, let $\mathbf{Q}_h \in \mathbb{R}^{T \times d_h}$ denote the slice of $\mathbf{Q}$ corresponding to the $h$-th head. Similarly, for each key-value group $g \in \mathcal{G}$, let $\mathbf{K}_g \in \mathbb{R}^{T \times d_h}$ and $\mathbf{V}_g \in \mathbb{R}^{T \times d_h}$ denote the slices of $\mathbf{K}$ and $\mathbf{V}$ corresponding to the $g$-th group. Note that each $\mathbf{K}_g$ and $\mathbf{V}_g$ has dimension $d_h$, matching the query head dimension.

Let $g(h) \in \mathcal{G}$ be the mapping function that assigns query head $h$ to its corresponding key-value group. The shared key and value matrices for the group associated with head $h$ are denoted as $\mathbf{K}_{g(h)}$ and $\mathbf{V}_{g(h)}$. 
In the standard FA mode, the output matrix for query head $h$ is computed as:
\begin{equation}
\mathbf{O}_h = \text{softmax}\left(\frac{\mathbf{Q}_h \mathbf{K}_{g(h)}^\top}{\sqrt{d_h}}\right)\mathbf{V}_{g(h)}.
\end{equation}

The outputs from all heads are concatenated along the feature dimension to form $\mathbf{O}=[ \mathbf{O}_{1},\mathbf{O}_{2},...,\mathbf{O}_{H} ] \in \mathbb{R}^{T \times d}$, which is then projected to produce the final layer output $\mathbf{Y} \in \mathbb{R}^{T \times d}$:
\begin{equation}
\mathbf{Y} = \mathbf{O} \mathbf{W}^O,
\end{equation}
where $\mathbf{W}^O \in \mathbb{R}^{d \times d}$ is the output projection matrix.

\subsection{Linear Attention with Gated DeltaNet (GDN)}
\label{subsec:gdn}
Although GQA reduces the KV cache footprint, its quadratic computational complexity still hinders the deployment of models in long-context scenarios, thereby motivating the development of linear attention mechanisms. Given that GDN~\cite{yang2025gated} has already been adopted in industrial-grade hybrid models, we likewise select GDN as the representative linear attention layer.

\paragraph{Standard Linear Attention.}
Standard linear attention~\cite{katharopoulos2020transformersrnnsfastautoregressive} maintains a matrix-valued recurrent state $\mathbf{S}_t \in \mathbb{R}^{d_h \times d_h}$ that accumulates key-value associations. Given head-wise input projections $\mathbf{q}_t, \mathbf{k}_t, \mathbf{v}_t \in \mathbb{R}^{d_h}$ at step $t$, the state update and output are defined as:
\begin{align}
\mathbf{S}_t &= \mathbf{S}_{t-1} + \mathbf{k}_t \mathbf{v}_t^\top, \\
\mathbf{y}_t &= \mathbf{S}_t^\top \mathbf{q}_t.
\end{align}
While computationally efficient, this formulation lacks a forgetting mechanism, causing the state to grow unbounded and leading to significant interference over long contexts.

\paragraph{Gated DeltaNet.}
To address these limitations, GDN interprets the state update as online gradient descent on a reconstruction loss, incorporating both a delta rule and a gating mechanism. Specifically, GDN introduces a scalar forget gate $\alpha_t \in (0, 1]$ and a learning rate $\beta_t$ to regulate memory retention. The state update rule is formulated as:
\begin{align}
\mathbf{S}_t = \alpha_t \left( \mathbf{I} - \beta_t \mathbf{k}_t \mathbf{k}_t^\top \right) \mathbf{S}_{t-1} + \beta_t \mathbf{k}_t \mathbf{v}_t^\top,
\end{align}
where $\mathbf{I} \in \mathbb{R}^{d_h \times d_h}$ is the identity matrix. The term $\left( \mathbf{I} - \beta_t \mathbf{k}_t \mathbf{k}_t^\top \right)$ acts as a projection that corrects the state based on the current key, while $\alpha_t$ serves as a data-dependent weight decay factor, controlling the lifespan of stored memories. The output is then computed as:
\begin{align}
\mathbf{y}_t = \mathbf{S}_t^\top \mathbf{q}_t.
\end{align}
This formulation allows GDN to maintain a constant-size hidden state $\mathbf{S}$ while dynamically balancing the retention of new information and the forgetting of outdated associations, thereby facilitating stable and efficient linear-time inference.

\section{Method}

We present the design of our head-wise hybrid architecture in three parts. Section~\ref{sec:head_selection} details an interpretability-based head selection algorithm that identifies critical heads to be preserved with full attention. Section~\ref{sec:head-wise-hybrid} then describes how FA and LA structures decompose input features and are organically fused at the head level. Finally, Section~\ref{sec:training} explains the transfer learning pipeline for efficiently converting a pretrained standard FA model into the proposed hybrid.

\subsection{Head Importance Estimation via Causal Intervention}
\label{sec:head_selection}
Deciding which heads should retain full attention---and which can be safely approximated by linear recurrence---is the decision on which our efficiency–quality trade-off hinges. Our procedure operates in three steps: (1) measure each head's direct causal effect on the target behavior via activation patching (\emph{receivers}); (2) trace upstream contributions via iterative path patching (\emph{senders}); and (3) fuse the per-head scores across multiple target capabilities into a single ranking. We describe each step below.

Observational diagnostics such as logit lens projections (Figure~\ref{fig:comm_headlogit}) or attention-pattern visualization can reveal head-level differences, but they show \emph{what} a head computes, not \emph{whether the model depends on it}: a head may attend strongly to a target token yet be overridden downstream, or attend diffusely but carry the critical signal in its output projection. Only causal intervention---replacing a head's output and observing the behavioral change---tests necessity directly, distinguishing load-bearing heads from correlated bystanders.

We therefore estimate, for each query head, its \emph{causal necessity} for the model's target capabilities, and retain FA only for the heads that prove causally indispensable. Building on the causal patching framework introduced in Section~\ref{sec:ap_prelim}, we describe the procedure below for an abstract set of target capabilities $\mathcal{C}$; its concrete instantiation (long-context retrieval and general ability) is deferred to Section~\ref{sec:experiments}.

\paragraph{Counterfactual construction and readout.}
For a capability $c \in \mathcal{C}$ we construct paired inputs (Section~\ref{sec:ap_prelim}) by replacing the correct answer in each prompt with a randomly generated counterfactual of the same type and token length (e.g., substituting one multi-digit number for another), producing a structure- and length-preserving edit that flips the target answer while holding the surrounding context fixed. This symmetric token replacement keeps every activation on the model's natural data distribution, avoiding the out-of-distribution artifacts of additive-noise corruption~\cite{zhang2024towardsbestpractices}. Since our target capabilities involve multi-token answers (e.g., multi-digit retrieval results), we extend the standard single-token logit difference (Eq.~\ref{eq:logit_diff_basic}) to a span-level readout. Let $\mathcal{A}$ denote the set of answer-token positions and let $a^{+}_j$, $a^{-}_j$ be the correct and counterfactual tokens at position $j \in \mathcal{A}$. We aggregate with an exponential decay factor $\lambda \in (0,1)$, where each position $j$ receives weight $\lambda^j$, emphasizing the earlier, more informative answer tokens:
\begin{equation}
    m(x) = \frac{1}{Z}\sum_{j \in \mathcal{A}} \lambda^{\,j}\big( z_j[a^{+}_j] - z_j[a^{-}_j] \big),
    \qquad Z = \sum_{j \in \mathcal{A}} \lambda^{\,j},
    \label{eq:logit_diff}
\end{equation}
where $z_j$ denotes the output logits at position $j$, and we use $\lambda = 0.9$ by default. We adopt the logit difference rather than probability or cross-entropy because it is approximately linear in the residual stream and monotone in the underlying capability, avoiding the softmax-saturation and probability measurement-floor effects that distort probability-based scores~\cite{zhang2024towardsbestpractices,nanda2023attributionpatching}.

\paragraph{Necessity via activation patching (receivers).}
We apply activation patching (Section~\ref{sec:ap_prelim}) at the granularity of individual heads, with a critical refinement: we freeze the downstream attention outputs to their clean values, so that the patched signal can propagate only through the residual stream (the running sum of all component outputs that accumulates across layers)---a \emph{direct-effect} measurement. Without this refinement, downstream heads could compensate for the corrupted signal, masking the true importance of the patched head. The head's importance, which we denote $\mathrm{IE}_{l,h}$, is the resulting drop in the readout, normalized to the clean--corrupt range:
\begin{equation}
    \mathrm{IE}_{l,h} = \frac{ m(x) - m\big(x;\ \mathbf{O}_{l,h}\!\leftarrow\!\mathbf{O}_{l,h}(x') \big) }{ m(x) - m(x') } \in [0,1],
    \label{eq:ie}
\end{equation}
where $m(x;\ \mathbf{O}_{l,h}\!\leftarrow\!\mathbf{O}_{l,h}(x'))$ denotes the readout when head $h$ at layer $l$ has its output replaced with the corrupted-run value while all other components remain at their clean state. The score is averaged over a small calibration set of counterfactual pairs drawn from the target capability (see Section~\ref{subsec:interp_anatomy} for the exact configuration). A value $\mathrm{IE}_{l,h}\!\approx\!0$ marks a head whose corruption leaves the behavior intact (dispensable, and safely approximated by linear recurrence), whereas $\mathrm{IE}_{l,h}\!\approx\!1$ marks a head whose corruption \emph{alone} collapses the capability. High-$\mathrm{IE}$ heads are the \emph{receivers}---heads that directly write the capability-critical signal into the residual stream. We denote the receiver importance of head $h$ as $\mathrm{IE}^{\text{recv}}_{h}$.

\paragraph{Upstream attribution via path patching (senders).}
A head can be causally important without writing to the output directly, by feeding a receiver. We trace one step back with path patching (Section~\ref{sec:ap_prelim}): for each candidate upstream head, we run the corrupted input and record the activations it sends to the receivers; we then run an otherwise-clean forward pass but substitute only those recorded activations at the receivers' inputs. The resulting normalized drop in the readout, which we denote $\mathrm{IE}^{\text{send}}_{h}$, measures the upstream head's contribution through the receiver pathway. We iterate, promoting the senders found in one round to receivers in the next, until new contributions vanish. For long-context retrieval, this converges in roughly two rounds---a \emph{shallow} circuit---so a small top-$K$ head set, rather than a full multi-stage circuit, already captures almost all of the causal signal.

\paragraph{Per-capability score and cross-capability fusion.}
For capability $c$ we combine the two roles by the larger of the receiver and sender effects, weighted by a task-consistency factor $\kappa^{(c)}_h$ equal to the fraction of sub-probes in which $h$ exceeds the importance threshold (i.e., $\mathrm{IE}_{l,h} \ge 0.01$) in its strongest role. This factor down-weights heads that achieve a high importance score on a single sub-probe but are negligible on the rest, favoring heads whose contribution is stable across tasks:
\begin{equation}
    s^{(c)}_h = \max\!\big(\mathrm{IE}^{\text{recv}}_{h},\, \mathrm{IE}^{\text{send}}_{h}\big)\cdot \kappa^{(c)}_h .
    \label{eq:per_cap_score}
\end{equation}
To preserve \emph{all} target capabilities jointly, we min--max normalize each $s^{(c)}$ to $[0,1]$ (the per-capability drops differ in scale), yielding $\hat{s}^{(c)}_h$, and fuse them by a weighted mean,
\begin{equation}
    S_h = \sum_{c \in \mathcal{C}} w_c\, \hat{s}^{(c)}_h, \qquad \sum_{c \in \mathcal{C}} w_c = 1,
    \label{eq:fusion}
\end{equation}
with equal weights by default, reflecting an equal prior over capabilities when no task-specific preference is given. We rank heads by $S_h$ and retain FA for the top-$K$ heads, assigning the remainder to LA, with $K$ set by the target FA budget.

\begin{figure}
    \centering
    \includegraphics[width=1\linewidth]{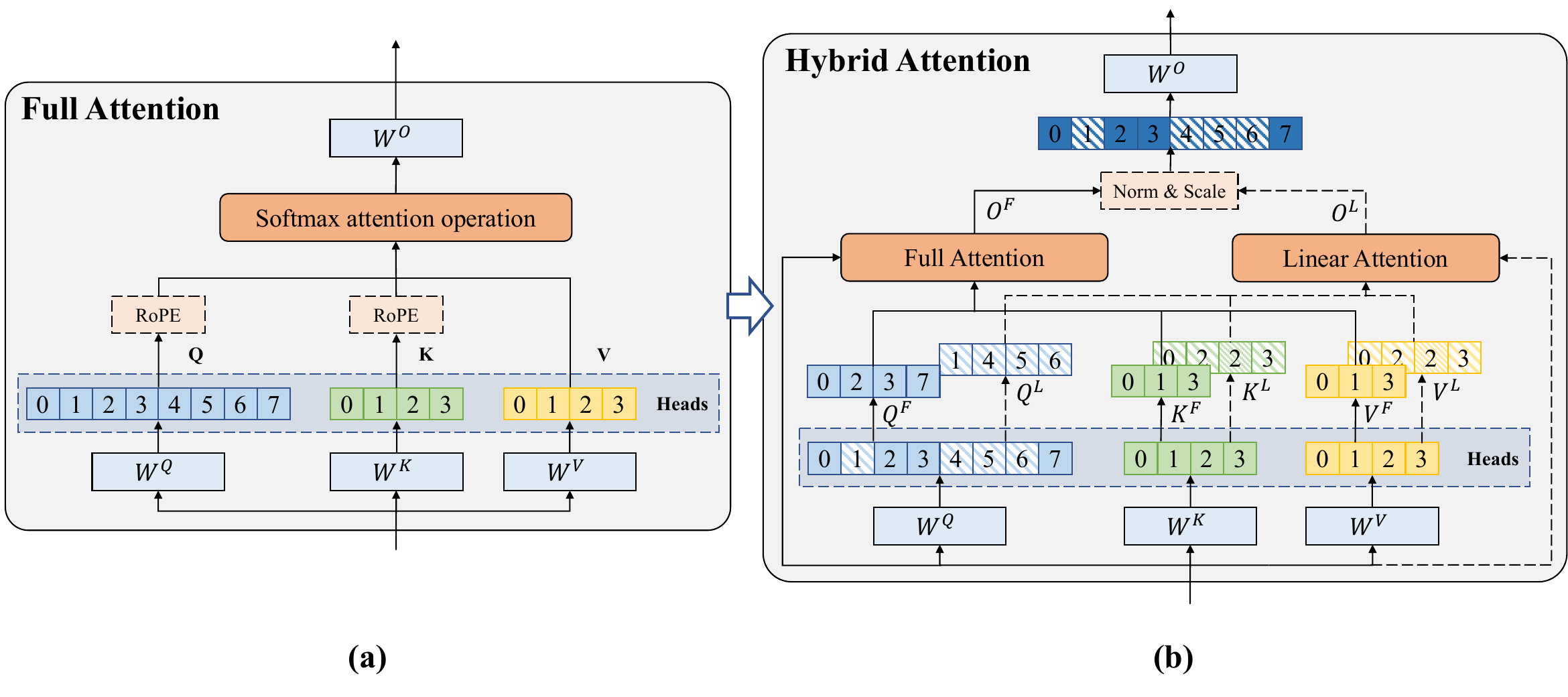}
    \caption{Comparison of standard FA and our proposed head-wise hybrid attention. A subset of heads retains the original FA branch, while the remaining heads employ GDN as a representative LA variant to reduce computational overhead.}%
    \label{fig:framework}
\end{figure}

\subsection{Head-wise Hybridization}
\label{sec:head-wise-hybrid}

FA excels at precise content retrieval through token-level interaction, yet its quadratic complexity constrains scalability to extremely long sequences. LA, by contrast, enables efficient long-context processing with $O(T)$ complexity, but its fixed-size recurrent state inevitably compresses historical information, leading to degraded recall fidelity over extended contexts. To harness the complementary strengths of both mechanisms, we propose a \textit{Head-wise Hybrid Attention} mechanism in Figure~\ref{fig:framework}. Instead of applying a uniform attention pattern across all heads or layers, we selectively assign each query head to either the FA branch or the LA (GDN) branch based on its functional importance. This fine-grained hybridization allows the model to retain critical reasoning capabilities in specific heads via FA, while leveraging the 
$O(T)$ efficiency and stable state evolution of GDN to facilitate effective context extension and reduce overall computational complexity.

\paragraph{Head Partitioning.}
Let $\mathcal{H} = \{1, \dots, H\}$ denote the set of all query head indices. We partition $\mathcal{H}$ into two disjoint subsets: $\mathcal{H}_F$ for heads assigned to the FA, and $\mathcal{H}_L$ for heads assigned to the GDN branch, such that $\mathcal{H}_F \cup \mathcal{H}_L = \mathcal{H}$ and $\mathcal{H}_F \cap \mathcal{H}_L = \emptyset$. The assignment is determined via an interpretability-driven selection process (detailed in Section~\ref{sec:head_selection}), which identifies heads crucial for precise context retrieval and complex reasoning.

\paragraph{Parallel Branch Computation.}
For each head $h \in \mathcal{H}$, the input projections $\mathbf{Q}_h, \mathbf{K}_{g(h)}, \mathbf{V}_{g(h)} \in \mathbb{R}^{T \times d_h}$ are computed as defined in Section~\ref{subsec:gqa}. These projections are then routed to their respective branches:
for $h \in \mathcal{H}_F$, the output $\mathbf{O}_h^F \in \mathbb{R}^{T \times d_h}$ is computed via standard softmax attention;
for $h \in \mathcal{H}_L$, the output $\mathbf{O}_h^L \in \mathbb{R}^{T \times d_h}$ is computed via the Gated DeltaNet recurrence.

\paragraph{Head-wise Scale-normalized Fusion.}
A fundamental asymmetry exists between FA and LA at the level of attention distributions. As NaLaFormer~\cite{meng2026normtimesdirectionrestoringmissingquery} reveals, FA's softmax exponential function naturally incorporates query vector magnitude into attention scores, producing sharp, low-entropy distributions concentrated on a few key tokens. LA's normalization operations, by contrast, cancel out query magnitude, yielding smoother, higher-entropy distributions. This divergence in attention patterns directly propagates to the output features: FA heads produce representations strongly modulated by query norm and dominated by a small set of high-weight tokens, while LA heads generate more uniform, query-norm-agnostic representations.  These distributional shifts may destabilize optimization when the two feature streams are naïvely concatenated.\footnote{We observe that when FA and LA are unevenly distributed, directly concatenating features leads to a significant degradation in capabilities in Table~\ref{tab:global_selection}.}
To mitigate this, we propose an independent normalization and head-wise modulation scheme that reconciles the distributional gap while preserving the structural integrity of the original attention heads.

Let $\mathbf{O}_h^F \in \mathbb{R}^{T \times d_h}$ and $\mathbf{O}_h^L \in \mathbb{R}^{T \times d_h}$ denote the raw outputs of the FA and GDN branches for head $h$, respectively. We first apply RMSNorm independently to each head's output to unify their feature scales:
\begin{equation}
    \hat{\mathbf{O}}_h = \text{Norm}(\mathbf{O}_h), \quad \text{where } \mathbf{O}_h = 
    \begin{cases} 
        \mathbf{O}_h^F & \text{if } h \in \mathcal{H}_F, \\
        \mathbf{O}_h^L & \text{if } h \in \mathcal{H}_L.
    \end{cases}
    \label{eq:norm}
\end{equation}

Crucially, we concatenate these normalized head outputs $\{\hat{\mathbf{O}}_1, \dots, \hat{\mathbf{O}}_H\}$ along the head dimension \textit{according to their original head indices} to form a tensor $\hat{\mathbf{O}} \in \mathbb{R}^{T \times H \times d_h}$. This index-preserving concatenation ensures that the positional and functional identity of each head is maintained within the global representation. 

To fully utilize the characteristics of these two attention mechanisms, we introduce a learnable head-wise scaling vector $\boldsymbol{\gamma} \in \mathbb{R}^{H}$. This vector assigns a scalar weight to each head based on its original index, allowing the model to adaptively recalibrate the contribution of each head regardless of its underlying attention type. The modulated tensor $\tilde{\mathbf{O}} \in \mathbb{R}^{T \times H \times d_h}$ is computed via broadcasting multiplication along the head dimension:
\begin{equation}
    \tilde{\mathbf{O}}_{:,h,:} = \gamma_h \cdot \hat{\mathbf{O}}_{:,h,:}, \quad \forall h \in [1, H].
    \label{eq:head-scale}
\end{equation}

Finally, $\tilde{\mathbf{O}}$ is reshaped into $\mathbb{R}^{T \times d}$ (where $d = H \cdot d_h$) and projected by the shared output matrix $\mathbf{W}^O$ to produce the final attention output. This design ensures that the distributional gaps between FA and GDN features are bridged before fusion, while preserving the distinct functional roles of individual heads through learnable scalars aligned with their original positions.

\begin{figure}
    \centering
    \includegraphics[width=1\linewidth]{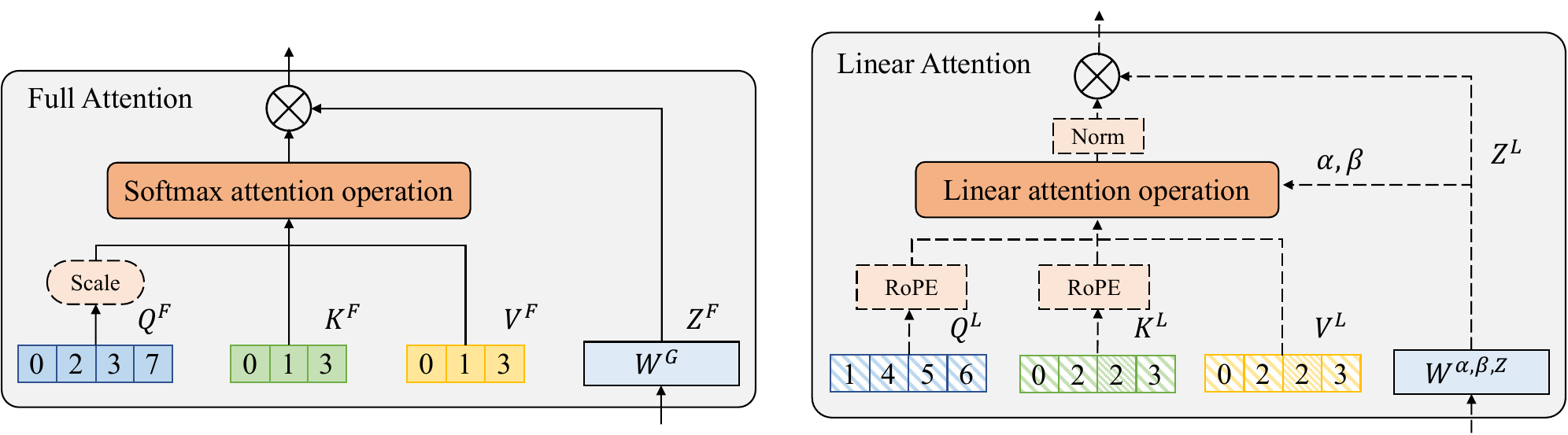}
    \caption{Details of FA and LA branches. We use GDN as a representative linear attention variant to illustrate the design. Specifically, we replace the original RoPE in the FA branch with a scaling mechanism, while applying RoPE to the LA (GDN) branch to enhance its positional awareness.}%
    \label{fig:attn_details}
\end{figure}

\paragraph{Branch-specific Refinements.}
The internal structural designs of the two attention types—such as positional encoding, gating mechanisms, and key-value head expansion—also exert a significant influence on head-wise hybridization. While HypeNet~\cite{chen2026hybridlinearattentionright} validates the effectiveness of these designs in the context of layer-wise mixing, we verify them under the setting of head-wise hybridization (Figure~\ref{fig:attn_details}). 

For the FA branch, we aim to enhance attention stability in long-context, especially extended-context scenarios. To this end, we depart from the standard Rotary Position Embedding (RoPE) protocol: following HypeNet~\cite{chen2026hybridlinearattentionright} and DroPE~\cite{gelberg2025extendingcontextpretrainedllms}, we remove RoPE and instead apply a log-scale coefficient to the query features after training, which effectively stabilizes the attention distribution. In addition, motivated by gated attention~\cite{qiu2025gatedattentionlargelanguage}, we introduce an auxiliary gate branch within the FA module. This design not only boosts the model's representational capacity but also effectively alleviates the ``attention sink'' phenomenon commonly observed in high-precision attention heads~\cite{xiao2024efficientstreaminglanguagemodels}.

For the GDN branch, we aim to strengthen its positional awareness within the inherent receptive field. Retaining the native short convolution and gating mechanism for local pattern recognition and stable state transitions, we explicitly integrate RoPE into the query and key projections to compensate for the limited positional sensitivity of linear recurrence, enabling precise capture of relative positional relationships within the perception window. Furthermore, to enhance representational capacity, we expand the number of key-value heads to match that of query heads—effectively transitioning from a GQA-like to an MHA-like configuration. Although described here in the context of GDN, these enhancements are directly transferable to other linear attention variants.

\subsection{Efficient Hybrid Transfer Learning}
\label{sec:training}

Transforming a pre-trained Transformer into a hybrid architecture offers a practical path to leveraging the strong performance of the original model while substantially reducing training overhead. The key challenge is that the newly introduced GDN branches start with fundamentally different computational dynamics than the FA heads they replace; direct fine-tuning from this mismatched initialization leads to optimization instability. Following prior work~\cite{li2025distillinghybridattentionmodels,chen2026hybridlinearattentionright}, we address this through a three-stage pipeline with a principled progression: first align local behavior (Stage 1: each layer's output should match the original FA layer), then align global semantics (Stage 2: the model's output distribution should match the teacher's), and finally adapt to the target domain (Stage 3: standard fine-tuning). Each stage builds on the stability achieved by the previous one.

\subsubsection{Stage 1: Parameter Migration and Layer-wise Output Alignment}

The first stage focuses on initializing the hybrid architecture and aligning the local behavior of each layer. After identifying the critical heads to remain as FA via our interpretability-driven selection mechanism, we initialize the remaining heads as GDN branches.

\paragraph{Parameter Migration Initialization.}
We employ a careful weight initialization strategy to minimize the distribution shift at the beginning of training:
\begin{itemize}
    \item \textbf{FA Branch with Gate Modulation:} For the heads retained as FA, we introduce a lightweight gate branch to allow for dynamic modulation during later stages. To encourage the gate to initially approximate an identity function (i.e., maintaining fidelity to the original FA output), we initialize the gate branch projection weights to near-zero and set the bias so that initial gate values are close to $1$. This guarantees that the FA branch starts with its pre-trained functionality intact, providing a stable anchor for the hybrid model.
    \item \textbf{GDN Branch Weight Reuse:} For the heads converted to GDN, we directly reuse the $Q, K, V$ projection weights from the original pre-trained FA layers. Since our model adopts GQA, where the number of KV heads is smaller than the number of Query heads, we handle the dimension mismatch by applying a channel-wise repeat technique. Specifically, the key-value projection matrices for the GDN branch are initialized by repeating the channels of the original pre-trained key-value weights to align with the required GDN internal dimensions. This ensures that the LA states initially capture similar key-value interactions as the original attention heads.
\end{itemize}

\paragraph{Layer-wise Hidden State Alignment.}
With the parameters initialized, we freeze the pre-trained backbone and optimize only the newly introduced hybrid attention layers. The objective is to align the output hidden states of each hybrid attention layer with those of the original pre-trained FA layer. We employ the Mean Squared Error (MSE) loss between the hidden state representations of the original FA layer ($\mathbf{H}_{FA}^{(l)}$) and the hybrid layer ($\mathbf{H}_{Hybrid}^{(l)}$) for each layer $l$:
\begin{equation}
    \mathcal{L}_{align} = \sum_{l=1}^{L} \| \mathbf{H}_{FA}^{(l)}(x) - \mathbf{H}_{Hybrid}^{(l)}(x) \|_2^2.
\end{equation}
This layer-wise MSE alignment ensures that the hybrid architecture locally mimics the feature space of the pre-trained model, preventing error accumulation in deeper layers and providing a robust foundation for subsequent global distillation.

\subsubsection{Stage 2: Global Logits Distillation}

In the second stage, we unfreeze the entire model and perform global knowledge distillation to align the final output distribution of the hybrid student model with the original pre-trained teacher model. Unlike Stage 1, which focuses on intermediate layer consistency, this stage targets the final logits to ensure global semantic coherence. We combine the KL divergence loss on the final logits with a standard cross-entropy loss on the ground truth labels:
\begin{equation}
    \mathcal{L}_{KD} = D_{KL}\left( P_{teacher}(\cdot|x) \parallel P_{student}(\cdot|x) \right),
\end{equation}
where $P_{teacher}$ and $P_{student}$ are the softmax probabilities of the teacher and student models, respectively. This stage allows the hybrid model to recover any minor performance degradation caused by the linear approximation and gate modulation, ensuring that the global linguistic capabilities of the pre-trained backbone are fully preserved.

\subsubsection{Stage 3: Long-Context Fine-tuning}

By this point, the hybrid architecture has been well-initialized and its output distribution aligned with that of the teacher model through the preceding stages. We therefore adopt standard supervised fine-tuning with the Next Token Prediction (NTP) objective in the final stage, using a longer context length than in the first two stages to consolidate long-context capabilities.
\begin{equation}
    \mathcal{L}_{NTP} = -\sum_{t} \log P_{student}(x_{t+1} | x_{1:t}).
\end{equation}

\section{Experiments}
\label{sec:experiments}

We conduct comprehensive experiments to evaluate HydraHead along four progressively refined questions: (1) \textbf{Does HydraHead outperform existing hybridization strategies?} (Section~\ref{subsec:exp_hybrid}): Controlled comparison against layer-wise, token-wise, and other head-wise alternatives under the same training pipeline. 
(2) \textbf{What makes it work?} (Section~\ref{subsec:ablation_structure}-\ref{subsec:feat_fusion}): Ablation of structural components and fusion mechanisms. 
(3) \textbf{Does interpretability-guided selection matter?} (Section~\ref{subsec:interpretability}-\ref{subsec:interp_anatomy}): Validation of causal selection vs. naive strategies, and analysis of head importance anatomy. 
(4) \textbf{How far can it scale?} (Section~\ref{subsec:optim_training}-\ref{subsec:scaling_up}): Training optimization and benchmarking against open-source models.

\subsection{Experimental Setup}

\subsubsection{Model, Training, and Head Selection Configuration}

\paragraph{Model and Data Configuration.} We conduct our primary experiments on the \textbf{Qwen3-1.7B} model~\cite{yang2025qwen3technicalreport}, which serves as the backbone for the initialization of our hybrid architecture. For training data, following the recent work~\cite{mamba,chen2026hybridlinearattentionright}, we utilize the \textbf{FineWeb-Edu} dataset~\cite{penedo2024finewebdatasetsdecantingweb}, a high-quality subset of the FineWeb corpus specifically filtered for educational content. This dataset is made available under the Open Data Commons Attribution License (ODC-By).

\paragraph{Model settings.} By default, we retain FA computation for 25\% of the heads, while the remaining 75\% use the GDN structure.

\paragraph{Head selection configuration.} We construct the calibration set for head importance estimation (Section~\ref{sec:head_selection}) from five NIAH sub-probes in the RULER benchmark (three single-key and two multi-key retrieval tasks), each comprising eight counterfactual pairs at 4K context length. All patching scores are computed in \texttt{fp16} precision. A head is deemed critical when its normalized importance $\lvert\mathrm{IE}_{l,h}\rvert \ge 0.01$; the per-head ranking is already stable from roughly six samples (validated in Section~\ref{subsec:interp_anatomy} and Appendix~\ref{apd:interp_details}).

\paragraph{Training details.} We train the hybrid models with the three-stage strategy as mentioned in Section~\ref{sec:training}. In our ablation studies, we follow the default settings of existing work \cite{chen2026hybridlinearattentionright}. Some training configurations within each stage are summarized in Table~\ref{tab:train_config}.

\begin{table}[]
    \centering
    \caption{Default training details of three training stages, which follow the HypeNet~\cite{chen2026hybridlinearattentionright}.}%
    \label{tab:train_config}
    \begin{tabular}{c|ccccc}
        \toprule
        Stage &  Batch Size & Context Length & Training Steps (Tokens) & LR scheduler & Learning Rate \\
        \midrule
        1 & 32 & 512 & 20K (0.3B) & Cosine & 1e-3 $\rightarrow$ 1e-5\\
        2 & 96 & 512 & 20K (1.0B) & Cosine & 1e-4 $\rightarrow$ 1e-5\\
        3 & 128 & 16384 & 500 (1.0B) & Constant & 1e-5 \\
        \bottomrule
    \end{tabular}

\end{table}

\subsubsection{Evaluation benchmarks}
To comprehensively assess the effectiveness of our method, we evaluate the model in two primary dimensions: \textbf{Long-Context Retrieval Capability} (Needle In A Haystack) and \textbf{General Reasoning Proficiency}.

\paragraph{Long-Context Retrieval (RULER).}
We employ the RULER benchmark~\cite{hsieh2024rulerwhatsrealcontext} to evaluate the model's ability to retrieve information from extended contexts. Specifically, we focus on two categories within RULER: \textit{Single-Key} (1/2/3) and \textit{Multi-Key} (1/2) retrieval tasks, which test the model's precision in locating single or multiple distinct pieces of information embedded in long sequences. We evaluate performance at five context lengths: 16K, 32K, 64K, 128K, and 256K tokens. Given that Qwen3-1.7B has a native context window of 32K, we categorize the evaluation results into two metrics:
\begin{itemize}
\item \textbf{Native Context Performance:} The average accuracy across 16K and 32K lengths, reflecting the model's baseline capability within its pre-trained window.
\item \textbf{Extended Context Performance:} The average accuracy across 64K, 128K, and 256K lengths, measuring the model's extrapolation ability and the effectiveness of our hybrid architecture in handling sequences beyond the native limit.
\end{itemize}

\paragraph{General Reasoning and Knowledge.}
To ensure that hybrid attention does not compromise the model's general reasoning capabilities in practical scenarios, we evaluate on a broad suite of benchmarks spanning two tiers: standard academic benchmarks commonly used in prior work, and more challenging evaluations widely adopted for assessing frontier large language models.
\begin{itemize}
\item \textbf{Basic Competency (Easy):} Following prior works~\cite{mamba2,zhao2026switchattentiondynamicfinegrained,chen2026hybridlinearattentionright}, we include six widely used datasets that assess basic language understanding and common sense reasoning: \textit{ARC-Challenge}  (norm score)~\cite{clark2018thinksolvedquestionanswering}, \textit{ARC-Easy}~\cite{clark2018thinksolvedquestionanswering}, \textit{HellaSwag} (norm score)~\cite{zellers2019hellaswag}, \textit{Winogrande}~\cite{sakaguchi2019winogrande}, \textit{PIQA}~\cite{bisk2019piqareasoningphysicalcommonsense}, and \textit{LAMBADA-OpenAI}~\cite{paperno2016lambadadatasetwordprediction}. These tasks serve as a sanity check for the model's foundational capabilities.
\item \textbf{Advanced Reasoning (Hard):} To probe deeper cognitive abilities, we evaluate on four challenging benchmarks requiring complex logical deduction, mathematical reasoning, and code generation: \textit{MMLU}~\cite{hendrycks2021measuringmassivemultitasklanguage}, \textit{GSM8K}~\cite{cobbe2021trainingverifierssolvemath}, \textit{MBPP}~\cite{austin2021programsynthesislargelanguage}, and \textit{BBH}~\cite{suzgun2022challengingbigbenchtaskschainofthought}. This tier critically tests whether the efficiency gains from our hybrid design come at the cost of high-level reasoning performance.
\end{itemize}

All evaluations are conducted using the lm-evaluation-harness framework~\cite{eval-harness}, adhering to its default configurations for each benchmark. This includes standard zero-shot or few-shot settings and greedy decoding as specified by the original protocols, unless otherwise noted.

\subsection{Hybrid Architecture Comparison}
\label{subsec:exp_hybrid}
We first systematically compare HydraHead against representative hybridization baselines spanning layer-wise, token-wise, and head-wise paradigms under the transfer learning protocol described in Section~\ref{sec:training}, to elucidate its distinct advantages.

\begin{table}[t]
    \centering
    \caption{Architecture comparison of various hybrid designs. $\text{Cache}$ denotes the relative key-value cache under a 16K context length, where the standard FA cache is normalized to 1.0. 
    $\text{Light.}$ denotes the lightning attention~\cite{qin2024variouslengthsconstantspeed}. 
    $^*$ and $^\ddagger$ indicate the hybrids are similar to HypeNet~\cite{chen2026hybridlinearattentionright} and Liger~\cite{lan2025liger}, respectively. $^\dagger$ indicates that an FA layer is inserted after every 3 LA layers. More details are provided in Appendix~\ref{apd: hybrid_arch}.}%
    \label{tab:arc_comparison}
    \begin{tabular}{lcccccccc}
        \toprule
        \multirow{2}{*}{Model} & \multirow{2}{*}{LA} & \multirow{2}{*}{Cache} & \multicolumn{2}{c}{RULER Single} & \multicolumn{2}{c}{RULER Multi-Key} & \multicolumn{2}{c}{General Reasoning} \\
        \cmidrule(lr){4-5} \cmidrule(lr){6-7} \cmidrule(lr){8-9}
         & & & (Native) & (Extended) & (Native) & (Extended) & (Hard) & (Easy) \\
        \midrule
        \multicolumn{9}{c}{Layer-wise Hybrid Transformers} \\
        \midrule
        FA \& LA$^*$ & Light. & 0.25  & 89.07 & 85.00 & 24.85 & 24.37 & 19.80 & 59.72 \\
        FA \& LA$^\dagger$ & Light. & 0.25  & 86.03 & 67.76 & 31.70 & 17.80 & 15.94 & 60.29\\
        FA \& LA & GDN & 0.25 & 47.63 & 7.02 & 22.50 & 5.53 & 16.44 & 58.18 \\
        LA-FA \& LA & GDN & 0.25 & 58.50 & 8.13 & 19.00 & 6.83 & 18.42 & 58.70 \\
        \midrule
        \multicolumn{9}{c}{Token-wise \& Head-wise Hybrid Transformers} \\
        \midrule
        Token-wise & GDN & 0.25 &  20.77 & 3.73 & 16.05 & 2.43 & \textbf{47.31} & 63.40\\
        Token-wise$^\ddagger$ & GDN & 0.25 & 96.03 & 58.78 & 26.20 & 11.03 & 36.66 & \textbf{63.37}\\
        Head-wise Mixing & GDN & 1.00 & 93.20 & 60.42 & \textbf{37.35} & 14.53 & 38.07 & 62.77\\
        HydraHead & GDN & 0.35 & \textbf{98.47} & \textbf{87.49} & 37.10 & \textbf{27.37} & 31.03 & 62.12\\
        \bottomrule
    \end{tabular}

\end{table}

\paragraph{Experimental Setup.}
We adopt GDN as the representative linear attention mechanism and construct baselines across three hybridization paradigms: layer-wise, token-wise, and head-wise.
For \textbf{layer-wise hybrids}, we follow~\cite{chen2026hybridlinearattentionright} to selectively retain FA in 25\% of the layers while employing GDN in the remainder, and additionally include a uniform variant that evenly distributes FA layers for comparison.
For \textbf{token-wise hybrids}, we consider two variants. The first combines 4K-context Sliding Window Attention (SWA) with GDN within every layer. The second adopts a layer-interleaved design inspired by Liger~\cite{lan2025liger}: 75\% of the layers use SWA (window size 128) combined with GDN, while the remaining 25\% combine full FA with GDN. 
For \textbf{head-wise mixing}, we implement a fusion-based variant inspired by~\cite{dong2024hymbahybridheadarchitecturesmall} that processes the input through parallel FA and GDN branches, each operating over the full set of heads, and fuses their outputs at the feature level.
For \textbf{head-wise selection}, existing per-head hybridization methods have focused on pairing FA with sparse attention~\cite{xiao2024duoattentionefficientlongcontextllm,tang2026elasticattentiontesttimeadaptive}. We use our HydraHead as the representative of this paradigm, retaining 25\% of heads as FA.
Detailed architectural specifications are provided in Appendix~\ref{apd: hybrid_arch}.

\paragraph{Results and Analysis.}
Table~\ref{tab:arc_comparison} presents a comprehensive study comparing our proposed head-wise hybrid architecture, HydraHead, against representative layer-wise, token-wise, and head-wise hybrids under identical training conditions and parameter budgets. Overall, we observe distinct performance trade-offs across architectures: while token-wise and head-wise hybrids demonstrate significant advantages in challenging general reasoning tasks, layer-wise hybrids generally exhibit superior capabilities in extending context length—with the notable exception of our method. Our HydraHead achieves the best overall balance, delivering state-of-the-art performance in long-context evaluations (particularly in context extrapolation) while maintaining robust general domain capabilities.

To understand these dynamics, we first examine the strength of token-wise and head-wise designs in complex reasoning. Compared to the layer-wise baseline, the other two mixing architectures (token-wise and head-wise) achieve substantially better performance in difficult general reasoning benchmarks, with gains exceeding 10\%. This suggests that fine-grained feature interaction within each layer is crucial for preserving the complex logical dependencies required for high-level reasoning. Layer-wise hybrid models, by uniformly substituting entire layers with LA, tend to disrupt these delicate attention patterns, leading to significant performance degradation in tasks requiring precise contextual understanding. However, this advantage in general reasoning traditionally comes at the cost of limited scalability in long-context scenarios, where naive token-wise and head-wise hybrid models struggle to maintain retrieval accuracy as the sequence length increases. In contrast, our model transcends this trade-off. By leveraging interpretability-guided head-wise selection, we preserve the critical global context awareness necessary for long-context extrapolation, thereby achieving superior performance in extended sequence tasks while maintaining a significant advantage in general-domain reasoning over layer-wise hybrids.

We note that the pure SWA+GDN token-wise variant (without FA-interleaved layers) exhibits notably inferior long-context performance compared to other variants, likely because the current transfer learning setup is insufficient for it to generalize to sequence lengths beyond 4× the training window.

\subsection{Structural Components Study}
\label{subsec:ablation_structure}

We conduct a systematic ablation study to dissect the contribution of each architectural component in our head-wise hybrid design, focusing on the interaction between the FA and GDN heads.

\paragraph{Experimental Setup: }
In this series of experiments, we adopt a \textbf{fixed head-wise allocation strategy} to isolate the impact of structural modules from the variability of dynamic routing. Specifically, for every transformer layer, we assign the first 75\% of attention heads to the GDN branch and the remaining 25\% to the FA branch (Figure~\ref{fig:head_distribution} (a)). This static configuration ensures a consistent computational budget across all variants, allowing for a fair comparison of how different internal mechanisms affect model expressiveness and stability.

\paragraph{Progressive Module Integration.}
We start from a minimal hybrid baseline and progressively introduce key modules to observe their individual and combined effects. The variants are defined as follows:
\begin{itemize}
\item \textbf{Base Hybrid}: A naive combination of FA and GDN heads without any specialized stabilization or feature alignment techniques. Note that the normalization of output (see Equation~\ref{eq:norm}) is used.
\item \textbf{+ FA NoPE \& Scale}: We remove positional embeddings (NoPE) for the FA branch and use position-related scaling instead. The scale hyperparameter is set to $500$.
\item \textbf{+ GDN RoPE}: We apply location encoding to the GDN branch to enhance content-aware capabilities. 
\item \textbf{+ FA Gate}: We incorporate a gating mechanism within the FA heads to dynamically modulate their contribution based on input context.
\item \textbf{+ GDN MHA}: Unlike FA, which uses GQA to reduce key-value cache, we use MHA in GDN to better utilize the multi-head mechanism to capture semantic associations.
\item \textbf{+ Query Decomposition}: Finally, we decompose the query projection matrices for FA and GDN heads. This avoids potential optimization bottlenecks in practice that may arise from numerical precision limitations~\cite{micikevicius2018mixedprecisiontraining} and implicit gradient conflicts between the two distinct heads.
\end{itemize}

\begin{table}[t]
    \centering
    \caption{Ablation study on structural components of the hybrid architecture under the \textbf{fixed head-wise allocation strategy}. Each module in each row is added based on the previous row.}%
    \label{tab:structure_update}
    \begin{tabular}{lcccccc}
        \toprule
        \multirow{2}{*}{Model} & \multicolumn{2}{c}{RULER Single} & \multicolumn{2}{c}{RULER Multi-Key} & \multicolumn{2}{c}{General Reasoning} \\
        \cmidrule(lr){2-3} \cmidrule(lr){4-5} \cmidrule(lr){6-7}
         & (Native) & (Extended) & (Native) & (Extended) & (Hard) & (Easy) \\
        \midrule
        Base Hybrid & 35.33 & 2.96 & 9.90 & 1.60 & \textbf{31.14} & 62.42 \\
        + FA NoPE \& Scale & 64.73 & 40.58 & \textbf{31.65} & 8.47 & 27.87 & 62.22\\
        + GDN RoPE & 72.80 & 56.38 & 24.00 & 14.00 & 26.36 & 62.09 \\
        + FA Gate & 76.07 & 60.89 & 31.05 & \textbf{15.07} & 26.98 & 62.60\\
        + GDN MHA & 79.97 & 54.27 & 31.35 & 14.17 & 29.66 & 62.42\\
        + Query Decomposition & \textbf{85.63} & \textbf{62.62} & 27.35 & 13.67 & 28.65 & \textbf{62.59}\\
        \bottomrule
    \end{tabular}

\end{table}

\begin{table}[t]
    \centering
    \caption{Query decomposition yields more pronounced gains under \textbf{interpretability-guided head selection} than under fixed allocation.}%
    \label{tab:split_q}
    \begin{tabular}{lcccccc}
        \toprule
        \multirow{2}{*}{Model} & \multicolumn{2}{c}{RULER Single} & \multicolumn{2}{c}{RULER Multi-Key} & \multicolumn{2}{c}{General Reasoning} \\
        \cmidrule(lr){2-3} \cmidrule(lr){4-5} \cmidrule(lr){6-7}
         & (Native) & (Extended) & (Native) & (Extended) & (Hard) & (Easy) \\
        \midrule
        Basic Model & 98.70 & 81.73 & 38.35 & 25.93 & 31.70 & 62.68\\
        w/o Query Decomposition & 91.30 & 79.43 & 46.85 & 25.77 & 31.79 & 62.78 \\
        \bottomrule
    \end{tabular}

\end{table}

\paragraph{Results and Analysis.}
Table~\ref{tab:structure_update} summarizes the performance variations resulting from the incremental addition of each module. We observe that replacing the positional encoding in the FA branch with scale modulation significantly improves the hybrid model’s long-context capabilities. In both Single-key and Multi-key tests, performance within the original context length increased by over 10\%, while extended context lengths improved from negligible gains to a substantial level. However, this modification reduces the positional awareness of the FA branch, leading to a decline of more than 3\% in general reasoning performance. Adopting a Multi-Head Attention (MHA) configuration for the GDN branch helps recover part of this lost general capability. The full configuration further decomposes the query projection matrices, resulting in the splitting of QKV projection matrices for both branches. This configuration achieves optimal performance on the RULER Single test while maintaining robust results in Multi-Key and general benchmarks. This advantage also holds consistently under the \textbf{interpretability-guided head selection}\footnote{It's denoted as Global Interpretability Screening in Section \ref{subsec:interpretability}.} setting (Table~\ref{tab:split_q}). Therefore, we denote this version model as \textbf{Basic Model}.

\subsection{Feature Fusion Designs}
\label{subsec:feat_fusion}

We systematically investigate feature fusion strategies based on a head-wise hybrid architecture. The core challenge lies in reconciling the different statistical properties of features generated using softmax and linear attention mechanisms—particularly differences in amplitude and variance. Directly concatenating these heterogeneous features often imposes significant optimization pressure on subsequent output projection layers, potentially leading to training instability or performance degradation.

\paragraph{Experimental Setup.} 
We conduct experiments under the \textbf{interpretability-guided head selection} strategy, where heads are selected based on their causal importance scores derived from mechanistic interpretability analysis. Unlike the fixed strategy, this global selection allows for heterogeneous layer-wise distributions; specific layers may exhibit extreme configurations, such as being composed entirely of GDN/FA heads, while others retain a higher concentration of GDN/FA heads. This setup creates a more complex and realistic optimization landscape, enabling us to rigorously assess the robustness of different fusion mechanisms in handling significant distributional shifts between attention branches.

\paragraph{Feature Fusion Variants.}
To systematically evaluate the impact of feature alignment strategies, we construct three variants based on the Basic Model, progressively introducing mechanisms to handle the distributional heterogeneity between FA and GDN branches:

\begin{itemize}
\item \textbf{Direct Concatenation (w/o Norm):} This baseline variant removes all normalization layers, directly concatenating the raw output features from FA and GDN heads. 

\item \textbf{Head-wise Scale Modulation:} Building upon a standard normalization step (e.g., RMSNorm) applied to each branch independently, this variant introduces learnable scalar coefficients (see Equation~\ref{eq:head-scale}). This allows the model to statically adjust the relative contribution magnitude of each head type during training. 

\item \textbf{Head-wise Gated Competition:} This variant replaces static scaling with a dynamic gating mechanism. A lightweight gate branch takes the original input sequence representation $\textbf{x}$ as input and produces head-wise modulation weights via a softmax activation, scaled by the total number of heads $H$ to preserve the overall feature magnitude. The final fusion is performed just like head-wise scale modulation. By enforcing a sum-to-one constraint across the FA and GDN components for each head position, this design introduces explicit competition, enabling context-aware selection between FA and GDN.
\end{itemize}

\begin{table}[t]
    \centering
    \caption{Comparison of feature fusion variants under the \textbf{interpretability-guided global selection}.}%
    \label{tab:global_selection}
    \begin{tabular}{lcccccc}
        \toprule
        \multirow{2}{*}{Model} & \multicolumn{2}{c}{RULER Single} & \multicolumn{2}{c}{RULER Multi-Key} & \multicolumn{2}{c}{General Reasoning} \\
        \cmidrule(lr){2-3} \cmidrule(lr){4-5} \cmidrule(lr){6-7}
         & (Native) & (Extended) & (Native) & (Extended) & (Hard) & (Easy) \\
        \midrule
        Basic Model & \textbf{98.70} & 81.73 & 38.35 & 25.93 & \textbf{31.70} & \textbf{62.68}\\
        w/o Norm & 85.07 & 71.36 & \textbf{41.90} & 23.93 & 30.62 & 62.58\\
        Head-wise Scale & 98.47 & \textbf{87.49} & 37.10 & \textbf{27.37} & 31.03 & 62.12\\
        Head-wise Gate & 95.80 & 67.16 & 41.85 & 26.87 & 28.24 & 61.90\\
        \bottomrule
    \end{tabular}

\end{table}

\paragraph{Results and Analysis.}
We present the comprehensive experimental results in Table~\ref{tab:global_selection}. Our analysis focuses on two key aspects: the necessity of feature normalization, and the trade-offs between static scaling and dynamic gating.

\textbf{Impact of Feature Normalization.}
The results reveal a consistent and substantial degradation across most evaluation dimensions. On RULER Single-key retrieval, removing normalization leads to a performance drop exceeding 10\% at both the original and extended context lengths, indicating that the feature distributions from heterogeneous attention heads are sufficiently misaligned to severely impair retrieval fidelity when naively combined. On challenging general reasoning benchmarks, a moderate but consistent degradation of approximately 1\% is observed. The sole exception is Multi-Key retrieval at the original context length, where direct concatenation yields a 3.55\% improvement—a counterintuitive result that warrants further investigation. One plausible explanation is that in the Multi-Key setting, the model must attend to multiple dispersed positions simultaneously, and the unnormalized feature mixing may inadvertently amplify certain cross-head interactions that facilitate multi-location tracking, at the cost of degraded single-location precision. Nevertheless, given the overwhelming degradation across all other metrics, this isolated gain does not justify removing normalization. We conclude that feature normalization is a critical design component for ensuring robust and universal performance across diverse head allocation strategies and evaluation scenarios.

\textbf{Scale Modulation vs. Gated Competition.}
We compare the two advanced fusion variants, Head-wise Scale Modulation and Head-wise Gated Competition, across the full evaluation suite. The results reveal a clear and consistent advantage for the Scale Modulation variant: it outperforms Gated Competition on nearly all metrics, with the sole exception of Multi-Key retrieval at the original context length. The most striking gain is observed in long-context extension scenarios, where Scale Modulation achieves a 20\% improvement over Gated Competition on Single-key retrieval tasks, demonstrating that static, learnable coefficients provide substantially more stable representations for modeling long-range dependencies under length extrapolation. Furthermore, compared to the Basic model, Scale Modulation delivers consistent improvements across all extended-context evaluations, confirming that the head-wise scaling mechanism generalizes robustly beyond the training distribution. The only setting where Gated Competition holds an edge—Multi-Key at the original context length—suggests that dynamic, context-aware gating may offer localized benefits when multiple retrieval targets are densely distributed within the training-length regime, but this advantage is not preserved under length expansion. Given the overwhelming and consistent superiority of Scale Modulation across both general reasoning and long-context benchmarks, we adopt Head-wise Scale Modulation as the \textbf{Default Model} for all subsequent experiments.

\begin{figure}[t]
\centering
\includegraphics[width=1\linewidth]{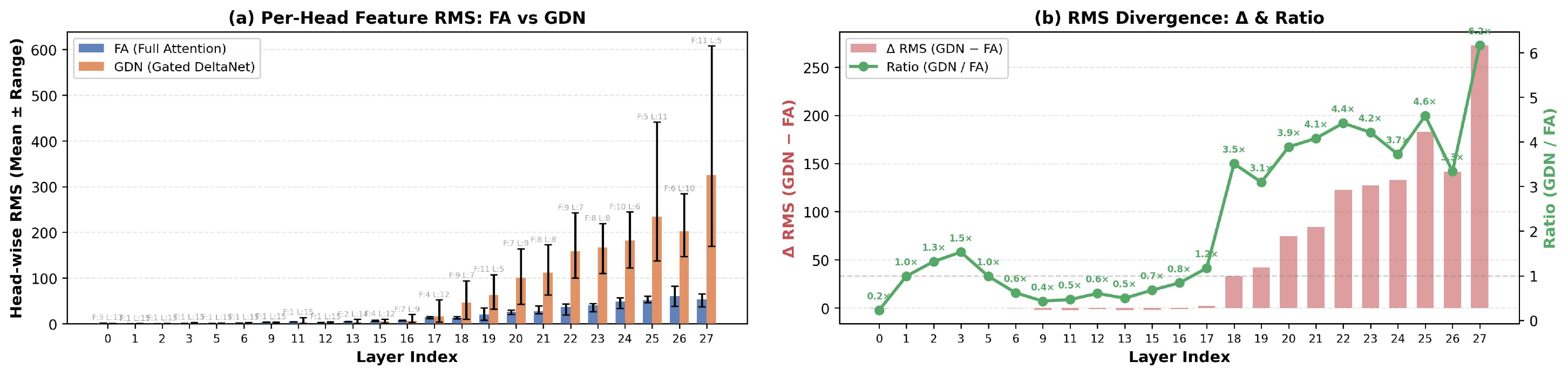}
\caption{Per-head feature RMS statistics of FA and GDN branches before normalization (Default Model). (a) Layer-wise RMS comparison with min--max range. (b) Divergence measured by $\Delta$ RMS and ratio.}
\label{fig:branch_rms}
\end{figure}

\paragraph{Feature Distribution Differences.} Figure~\ref{fig:branch_rms} reports the per-head feature RMS statistics of the FA and GDN branches prior to head-wise fusion. In shallow layers (0--17), the two branches exhibit comparable magnitudes (GDN/FA ratio within [0.2,1.5]), partly because shallow layers are dominated by GDN heads (Figure~\ref{fig:head_distribution}(e)), leaving few FA heads as a contrasting reference. In deeper layers (18--27), however, a pronounced scale mismatch emerges, with GDN RMS reaching up to 6.2× that of FA at layer 27. This divergence stems from the norm-awareness asymmetry between the two paradigms: softmax in FA constrains output magnitudes via exponential normalization, whereas linear attention in GDN lacks such regularization. The observed mismatch motivates independent RMSNorm for each branch before fusion, preventing either branch from dominating the combined representation due to scale disparities alone.

\subsection{Effectiveness of Interpretability Screening}
\label{subsec:interpretability}

To validate the hypothesis that functional heterogeneity among attention heads drives the efficacy of hybrid architectures, we investigate the impact of head selection strategies. This section aims to demonstrate that interpretability-guided selection, which prioritizes heads responsible for precise retrieval and factual recall, significantly outperforms naive distribution strategies. By aligning the architectural inductive bias with the intrinsic functional specialization of pre-trained heads, we aim to maximize the synergy between the high-precision FA branch and the efficient GDN branch.

\textbf{Experimental Setup.}
We compare five distinct head allocation strategies, all maintaining a global fixed ratio of 3:1 between GDN and FA heads to ensure comparable computational budgets. The baselines include: 
\begin{itemize}
    \item \textbf{(a) Fixed Allocation}, which uniformly assigns a fixed fraction of heads to FA in every layer.
    \item  \textbf{(b) Layer-wise Random Selection}, which randomly selects FA heads within each layer.
    \item \textbf{(c) Global Random Selection}, which randomly selects FA heads across the entire network.
\end{itemize}

\begin{figure}
    \centering
    \includegraphics[width=1\linewidth]{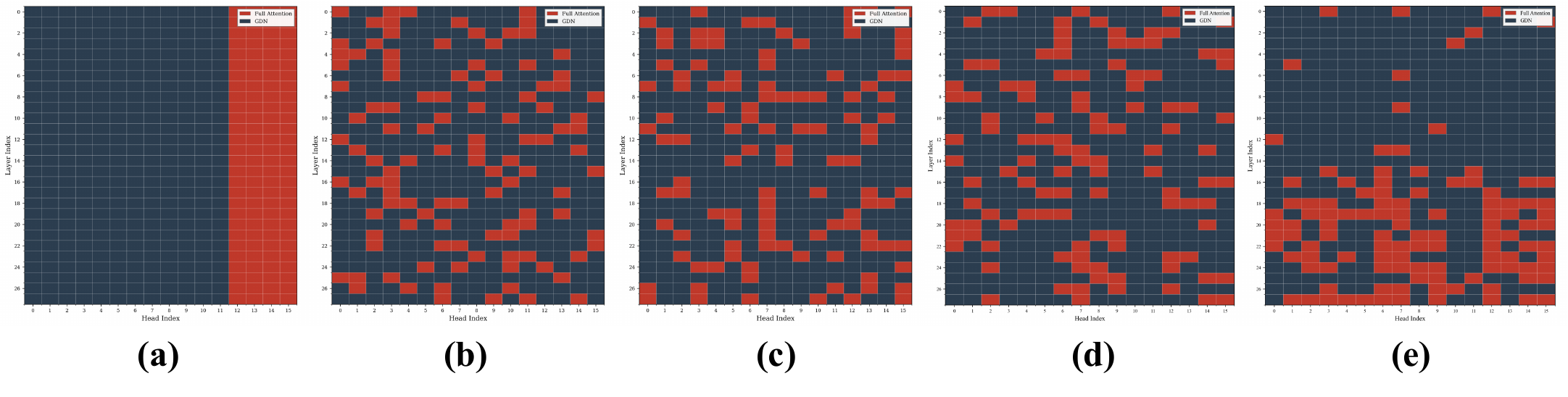}
    \caption{Comparison of distribution results of 5 head screening strategies: (a) Fixed Allocation; (b) Layer-wise Random Selection; (c) Global Random Selection; (d) Layer-wise Interpretability Screening; (e) Global Interpretability Screening.}%
    \label{fig:head_distribution}
\end{figure}

As our proposed methods, we evaluate two interpretability-driven variants based on causal importance scores: 
\begin{itemize}
    \item \textbf{(d) Layer-wise Interpretability Screening}, which selects the top-ranked heads within each layer according to their local importance.
    \item \textbf{(e) Global Interpretability Screening}, which selects the most critical heads across all layers regardless of their positional distribution.
\end{itemize}
Figure~\ref{fig:head_distribution} visualizes the resulting head distributions for these five strategies, highlighting the non-uniform, layer-skewed patterns emerging from the global interpretability-based approach. 

\begin{table}[t]
    \centering
    \caption{Comparison of different head selection strategies. All experiments are conducted based on \textbf{Basic Model}.}%
    \label{tab:head_selection}
    \begin{tabular}{lcccccc}
        \toprule
        \multirow{2}{*}{Model} & \multicolumn{2}{c}{RULER Single} & \multicolumn{2}{c}{RULER Multi-Key} & \multicolumn{2}{c}{General Reasoning} \\
        \cmidrule(lr){2-3} \cmidrule(lr){4-5} \cmidrule(lr){6-7}
         & (Native) & (Extended) & (Native) & (Extended) & (Hard) & (Easy) \\
        \midrule
        (a) Fixed & 85.63 & 62.62 & 27.35 & 13.67 & 28.65 & 62.59\\
        (b) Layer-Rand  & 87.07 & 65.33 & 27.70 & 17.37 & 27.91 & \textbf{62.95}\\
        (c) Global-Rand & 59.40 & 32.96 & 16.55 & 5.33 & 26.19 & 62.40\\
        \midrule
        (d) Layer-Interp & 91.13 & 65.22 & 35.10 & 15.97 & \textbf{31.93} & 62.15\\
        (e) Global-Interp & \textbf{98.70} & \textbf{81.73} & \textbf{38.35} & \textbf{25.93} & 31.70 & 62.68\\
        \bottomrule
    \end{tabular}

\end{table}

\paragraph{Results and Analysis.}
The comparative results across the five selection strategies are summarized in Table~\ref{tab:head_selection}. We have the following findings:

\textbf{Baseline Stability and Global Random Degradation.}
We observe that fixed allocation and layer-wise random selection yield nearly identical performance across all evaluation dimensions. This suggests that a uniform distribution of FA heads provides a stable baseline, and random variations within layers do not significantly alter the model's functional capacity. In contrast, global random selection leads to noticeable performance degradation, whether in single-key or multi-key retrieval tasks. This indicates that disrupting the layer-specific structural balance without functional guidance harms the model's ability to coordinate information flow across depths.

\textbf{Efficacy of Layer-wise Interpretability Screening.} Layer-wise interpretability screening demonstrates significant gains across almost all evaluation benchmarks, validating our hypothesis that preserving functionally critical heads (e.g., retrieval heads) in the FA branch is more effective than naive uniform allocation. By prioritizing heads with higher causal importance within each layer, this strategy ensures that local computational resources are allocated to the most impactful attention patterns.

\textbf{Unlocking Potential via Global Interpretability Screening.}
Global interpretability screening achieves the best overall performance, further unlocking the model's intrinsic potential. By selecting the most causally important heads across the entire network, this strategy allows for a non-uniform, optimized distribution where FA resources are concentrated in layers that contribute most to reasoning and recall. As visualized in Figure~\ref{fig:head_distribution}, this global approach results in a sparse but highly effective allocation pattern, proving that functional heterogeneity is a key driver for efficient hybrid attention design.

\textbf{Why interpretability outperforms naive strategies.}
The advantage of interpretability-guided selection has a structural explanation. Prior work has established that attention heads specialize in distinct functional roles~\cite{voita2019analyzingmultiheadselfattentionspecialized} and vary dramatically in importance---many can be pruned with minimal loss, while a few are indispensable~\cite{michel2019sixteenheadsreallybetter}. Our localization results (Section~\ref{subsec:interp_anatomy}) confirm this for long-context retrieval: only $\approx$6.5\% of heads are causally critical, and they are scattered across layers rather than concentrated in a few. Fixed and random allocation strategies are blind to this structure: they either waste FA budget on heads that do not need it (fixed assigns FA uniformly, regardless of function) or risk converting away critical heads entirely (global random can leave entire layers without FA, as the performance collapse in Table~\ref{tab:head_selection}(c) shows). Interpretability screening, by contrast, allocates every FA slot to a head that demonstrably carries retrieval signal, which is why it dominates at all ratios---and increasingly so at aggressive ratios (Table~\ref{tab:comp_ratio}), where the margin for wasted budget shrinks.

\subsection{Impact of Higher Hybrid Ratios}
\label{subsec:high_ratio}

\begin{figure}
    \centering
    \includegraphics[width=0.7\linewidth]{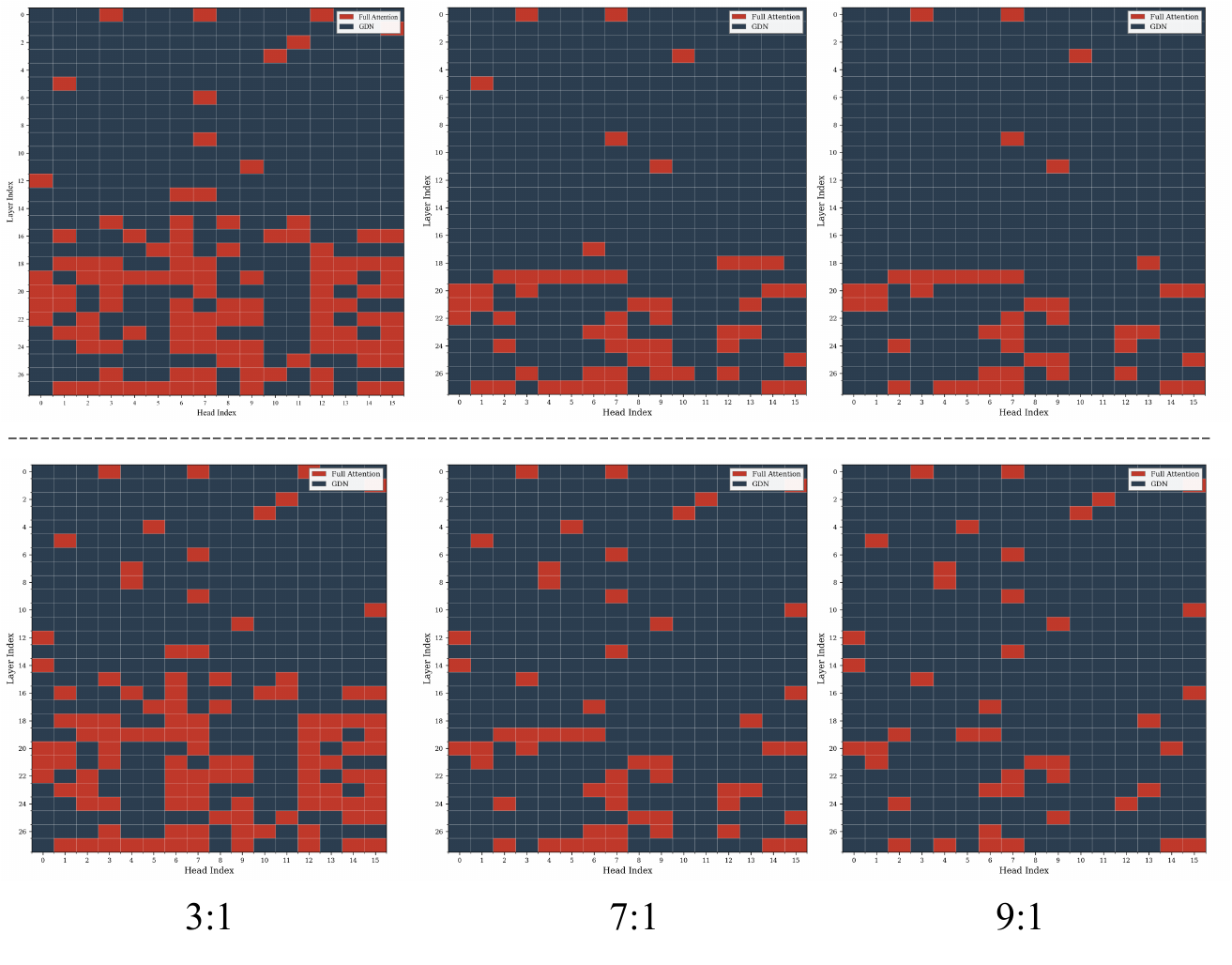}
    \caption{Comparison of head selection distributions under varying GDN-to-FA ratios. The top row illustrates the distribution obtained by direct screening based on head interpretability scores. The bottom row shows the distribution with an additional constraint that ensures at least one FA head is retained in each layer, while still prioritizing high-scoring heads.}%
    \label{fig:var_dist}
\end{figure}

We investigate the performance trends under higher GDN-to-FA mixing ratios to further explore the efficiency potential of our hybrid architecture. While a 3:1 ratio provides a balanced trade-off, increasing the proportion of GDN can significantly reduce memory footprint and computational overhead.

\paragraph{Experimental Setup.}
We evaluate three mixing ratios, extending from the baseline 3:1 to more aggressive configurations of 7:1 and 9:1. For each ratio, we compare two head selection strategies:
\begin{itemize}
\item \textbf{Global Interpretability Screening:} Heads are selected strictly based on their global causal importance scores, regardless of their layer distribution. This serves as the baseline to test the raw efficacy of interpretability-guided selection.
\item \textbf{Constrained Global Screening:} Recognizing that strict global ranking at high sparsity levels may concentrate FA heads in only a few layers—leaving most layers with pure GDN and potentially disrupting hierarchical information flow—we introduce a structural constraint. This strategy ensures that at least one FA head is retained in every layer, while the remaining FA budget is allocated to the highest-scoring heads globally.
\end{itemize}
This comparison allows us to disentangle the benefits of functional importance from the necessity of structural coverage across network depth. Figure~\ref{fig:var_dist} visualizes the resulting head distributions under both strategies. As the GDN-to-FA ratio increases, the proportion of layers composed entirely of GDN heads rises steadily, reaching nearly 50\% in the extreme 9:1 configuration.

\begin{table}[t]
    \centering
    \caption{Performance comparison under varying GDN-to-FA ratios. We evaluate two selection strategies: \textit{Global-Interp} (direct screening by interpretability scores) and \textit{Global-Interp-C} (with a constraint of at least one FA head per layer). All experiments are conducted based on \textbf{Basic Model}.}%
    \label{tab:comp_ratio}
    \begin{tabular}{lccccccc}
        \toprule
        \multirow{2}{*}{Model} & \multirow{2}{*}{Ratio} & \multicolumn{2}{c}{RULER Single} & \multicolumn{2}{c}{RULER Multi-Key} & \multicolumn{2}{c}{General Reasoning} \\
        \cmidrule(lr){3-4} \cmidrule(lr){5-6} \cmidrule(lr){7-8}
        &  & (Native) & (Extended) & (Native) & (Extended) & (Hard) & (Easy) \\
        \midrule
        Global-Interp & 3:1 & \textbf{98.70} & 81.73 & \textbf{38.35} & \textbf{25.93} & 31.70 & 62.68\\
        Global-Interp-C & 3:1 & 96.27 & \textbf{83.91} & 36.45 & 25.20 & \textbf{31.83} & \textbf{62.73}\\
        \midrule
        Global-Interp & 7:1 & 81.40 & 58.07 & 17.90 & 18.67 & 27.75 & 62.63\\
        Global-Interp-C & 7:1 & \textbf{88.70} & \textbf{81.04} & \textbf{26.70} & \textbf{22.93} & \textbf{29.46} & \textbf{62.65}\\
        \midrule
        Global-Interp & 9:1 & 89.12 & 60.58 & 18.65 & 14.50 & 26.04 & 62.65\\
        Global-Interp-C & 9:1 & \textbf{90.10} & \textbf{72.69} & \textbf{21.05} & \textbf{18.63} & \textbf{28.92} & \textbf{62.68}\\
        \bottomrule
    \end{tabular}

\end{table}

\begin{table}[t]
    \centering
    \caption{Comparison of layer-wise hybrids with a 3:1 LA:FA ratio (marked with $^*$ in Table~\ref{tab:arc_comparison}) against our \textbf{Basic Model} with a 7:1 ratio. Our model performs \textit{Global-Interp-C} head selection strategy.}%
    \label{tab:comp_7-1_to_3-1}
    \begin{tabular}{lc|cccccc|cc}
        \toprule
        \multirow{2}{*}{Model} & \multirow{2}{*}{Ratio} & \multicolumn{5}{c}{RULER Single \& Multi-Key} & \multirow{2}{*}{Avg} & \multicolumn{2}{|c}{General Reasoning} \\
        \cmidrule(lr){3-7} \cmidrule(lr){9-10} 
        &  & 16K & 32K & 64K & 128K & 256K &  & (Hard) & (Easy) \\
        \midrule
        Layer-wise hybrid & 3:1 & \textbf{60.52} & 53.40 & \textbf{60.79} & \textbf{53.59} & \textbf{49.69} & \textbf{55.59} & 19.80 & 59.72 \\
        Global-Interp-C & 7:1 & 59.55 & \textbf{55.85} & 57.35 & 51.04 & 47.59 & 54.27 & \textbf{29.46} & \textbf{62.65}\\
        \bottomrule
    \end{tabular}

\end{table}

\paragraph{Results and Analysis.}
Table~\ref{tab:comp_ratio} summarizes the performance results across different hybrid ratios. We can summarize the findings as follows:

\textbf{Performance Degradation and Structural Resilience.}
As the GDN-to-FA ratio increases, we observe a consistent decline in general reasoning capabilities, with complex reasoning tasks being particularly sensitive to the reduction of FA heads. In contrast, long-context performance exhibits a fluctuating downward trend. This divergence suggests that while LA can effectively handle local dependencies and simple retrieval, the loss of global receptive fields in FA heads disproportionately impacts multi-step logical deduction. The Global-Interp-C strategy, which ensures at least one FA head per layer, demonstrates significant advantages at higher sparsity levels (7:1 and 9:1). By preventing the complete collapse of FA capacity in certain layers, this constraint maintains the hierarchical information flow essential for deep networks. At lower ratios (3:1), while its impact on long-context tasks varies, it consistently yields improvements in general reasoning, indicating that structural diversity across layers is beneficial even when the overall FA budget is sufficient.

\textbf{Sparsity Limits and Practical Effectiveness.}
A direct comparison between the constrained 9:1 configuration and the standard 3:1 baseline reveals the current limitations of our hybrid architecture under extreme sparsity. While the model retains over 80\% of its performance on simple long-context tasks, it suffers from significant degradation in more complex scenarios. Specifically, accuracy drops to approximately 60\% in multi-key retrieval settings, and general reasoning capabilities on challenging benchmarks decline by about 3\%. These results indicate that while the layer-wise constraint effectively prevents structural collapse, the severe reduction in FA heads still compromises the model's ability to handle intricate dependencies and complex logical patterns, suggesting that further optimization is required to bridge the performance gap in extreme sparse regimes. Despite these limitations at extreme sparsity, our model retains a substantial advantage over layer-wise hybrids in practical regimes. At a 7:1 ratio, we achieve long-context performance on RULER (16K--256K) broadly comparable to the layer-wise hybrid at 3:1, with per-length differences within ±3\% (Table~\ref{tab:comp_7-1_to_3-1}). Meanwhile, it demonstrates a substantial advantage in general reasoning (+9.66\% on Hard, +2.93\% on Easy), underscoring the effectiveness of head-wise hybridization in preserving strong general-domain capabilities while enabling aggressive LA:FA ratios for long-context extension.

\subsection{Anatomy of Interpretability-Guided Head Selection}
\label{subsec:interp_anatomy}

Tables~\ref{tab:head_selection} and~\ref{tab:comp_ratio} establish \emph{that} interpretability-guided screening works; this section explains \emph{why}. We analyze the head-importance scores of Equation~\eqref{eq:fusion} computed on the pretrained dense Qwen3-1.7B checkpoint (28 layers $\times$ 16 heads $=448$ query heads) using the long-context retrieval probe, and show that (i) the score is a stable causal measurement, (ii) retrieval is localized to a sparse set of heads that does \emph{not} respect layer boundaries---the empirical foundation of head-level hybridization---and (iii) the score is causally faithful under knockout. A complementary analysis---the representational diversity of heads within a layer---is deferred to Appendix~\ref{apd:interp_diversity}.

\paragraph{Experimental Setup.}
We instantiate the procedure of Section~\ref{sec:head_selection} with long-context retrieval as the calibration capability, realized by five Needle-in-a-Haystack (NIAH) sub-probes from RULER (three single-key, two multi-key) that combine activation patching (receivers) with iterative path patching (senders). We write $\lvert\mathrm{drop}\rvert$ for the per-head normalized importance $\mathrm{IE}_{l,h}$ of Equation~\eqref{eq:ie} and, following common practice, call a head \emph{critical} when $\lvert\mathrm{drop}\rvert \ge 0.01$. Unless noted, importance is taken as the union over the five sub-probes, computed on a small calibration set.\footnote{In our experiments, the calibration set consists of a few samples (eight) localized at short context and scored in \texttt{fp16}; the per-head ranking is already stable from roughly six samples (see the stability analysis below and Appendix~\ref{apd:interp_details}).} General ability is the second capability fused under Equation~\eqref{eq:fusion}; its localization protocol is analogous and omitted for brevity.

\begin{figure}[t]
    \centering
    \includegraphics[width=0.32\linewidth]{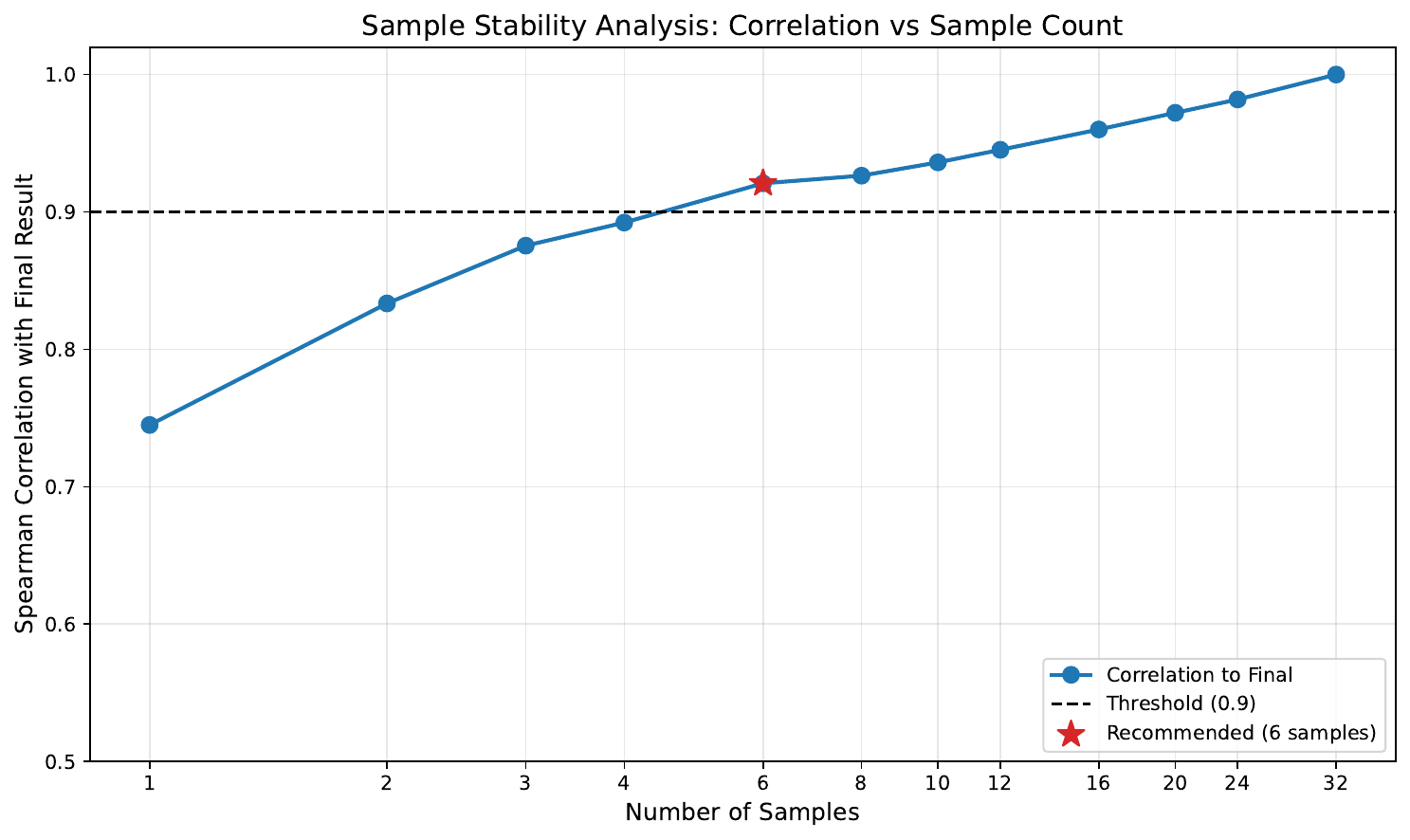}\hfill
    \includegraphics[width=0.32\linewidth]{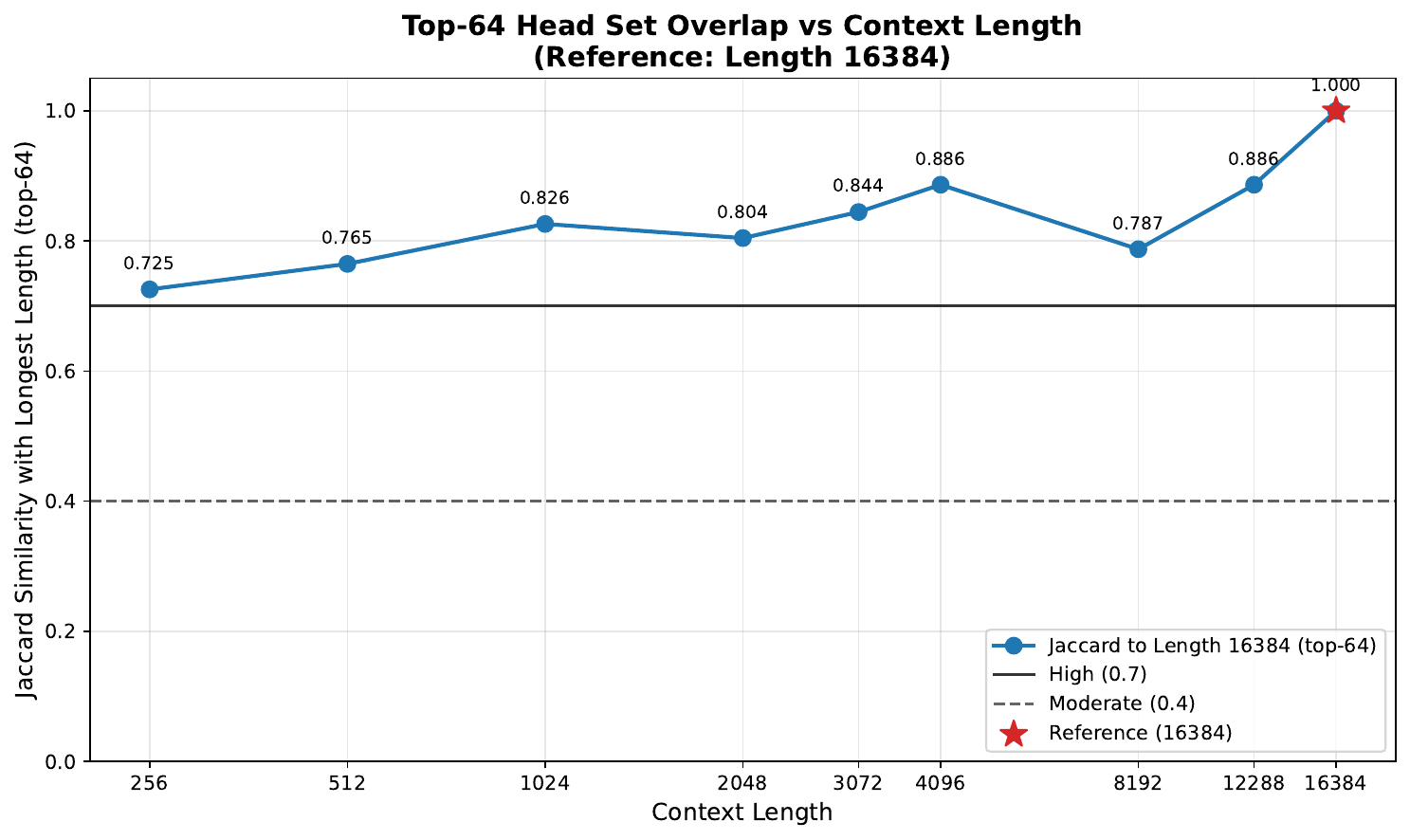}\hfill
    \includegraphics[width=0.32\linewidth]{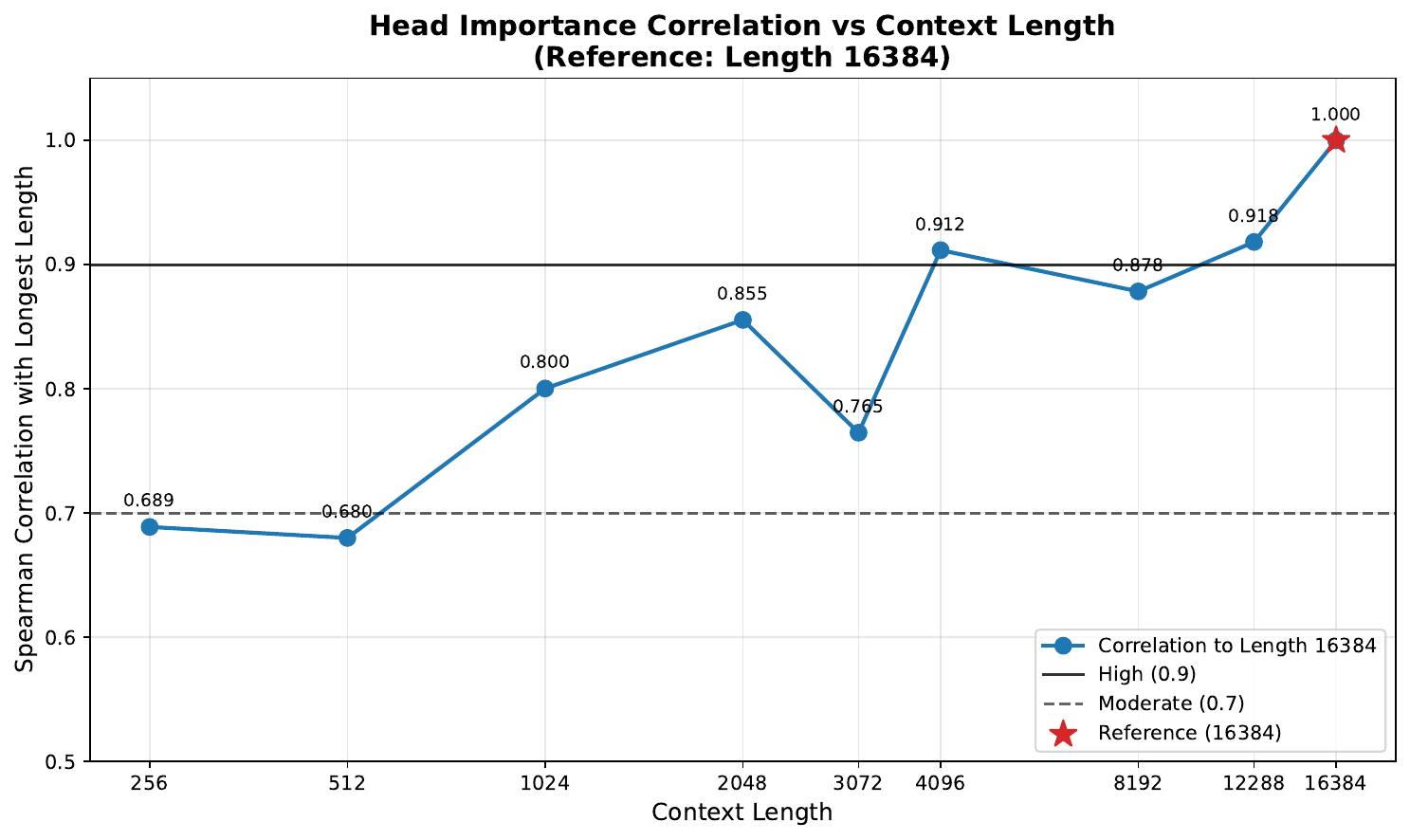}
    \caption{Stability and robustness of the head-importance ranking on Qwen3-1.7B. \textbf{(Left)} Sample stability: Spearman correlation between the rank from $k$ calibration samples and the $32$-sample reference (crosses the $0.9$ threshold at $k=6$, $\rho\approx0.921$, and saturates by $k=12$). \textbf{(Middle, Right)} Robustness to the localization context length, measured against the $16$k reference: top-$32$ Jaccard overlap of the critical-head set, and Spearman rank correlation. The head \emph{set} is well preserved across both samples and lengths, while exact ranks drift at intermediate lengths---justifying localization from a few short-context samples and transfer to long context.}%
    \label{fig:interp_validity}
\end{figure}

\paragraph{The importance score is a stable measurement.}
A selection rule is only as trustworthy as the measurement under it. Figure~\ref{fig:interp_validity} (left) reports the Spearman rank correlation between the per-head ranking computed from a growing number of calibration samples and a larger-sample reference: the ranking stabilizes after only a handful of samples and is essentially fixed thereafter, the empirical basis for the ``a few forward passes'' claim in Section~\ref{sec:head_selection}. The selection is also robust to the context length at which it is computed (Figure~\ref{fig:interp_validity}, middle and right): over a wide range of context lengths, the top-ranked \emph{set} is well preserved and converges toward the long-context reference, even though absolute ranks drift at intermediate lengths. Across the sub-probes, the critical-head sets overlap substantially, with number-needle probes agreeing most strongly and \texttt{uuid}-needle probes diverging; we therefore take the union across sub-probes, which is the conservative choice. The selection is thus stable at the level of the head \emph{set}, which is what we rely on, while individual ranks are noisier---consistent with the redundancy structure discussed in Appendix~\ref{apd:interp}.

\begin{figure}[t]
    \centering
    \begin{subfigure}[b]{0.20\linewidth}
        \centering
        \includegraphics[width=\linewidth]{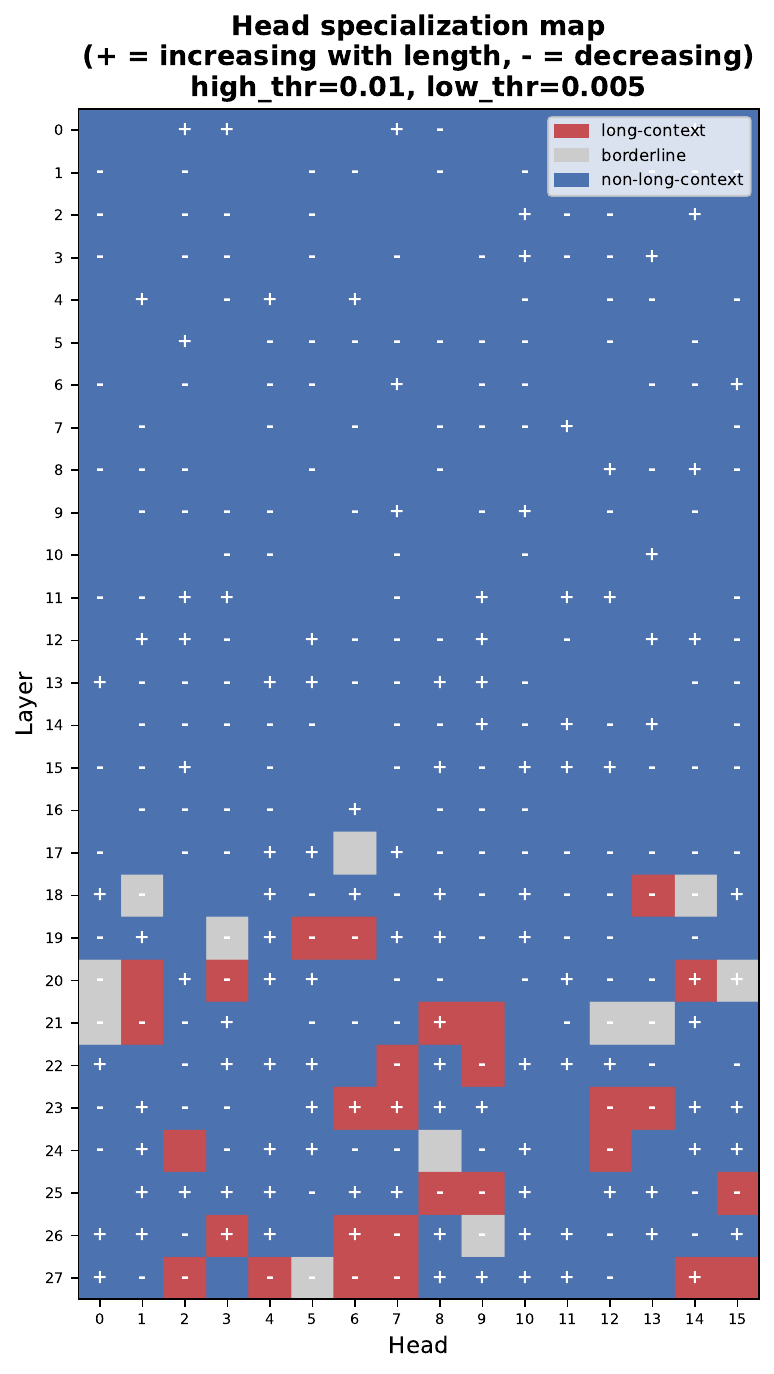}
        \caption{}%
        \label{fig:interp_granularity}
    \end{subfigure}
    \hfill
    \begin{subfigure}[b]{0.77\linewidth}
        \centering
        \includegraphics[width=\linewidth]{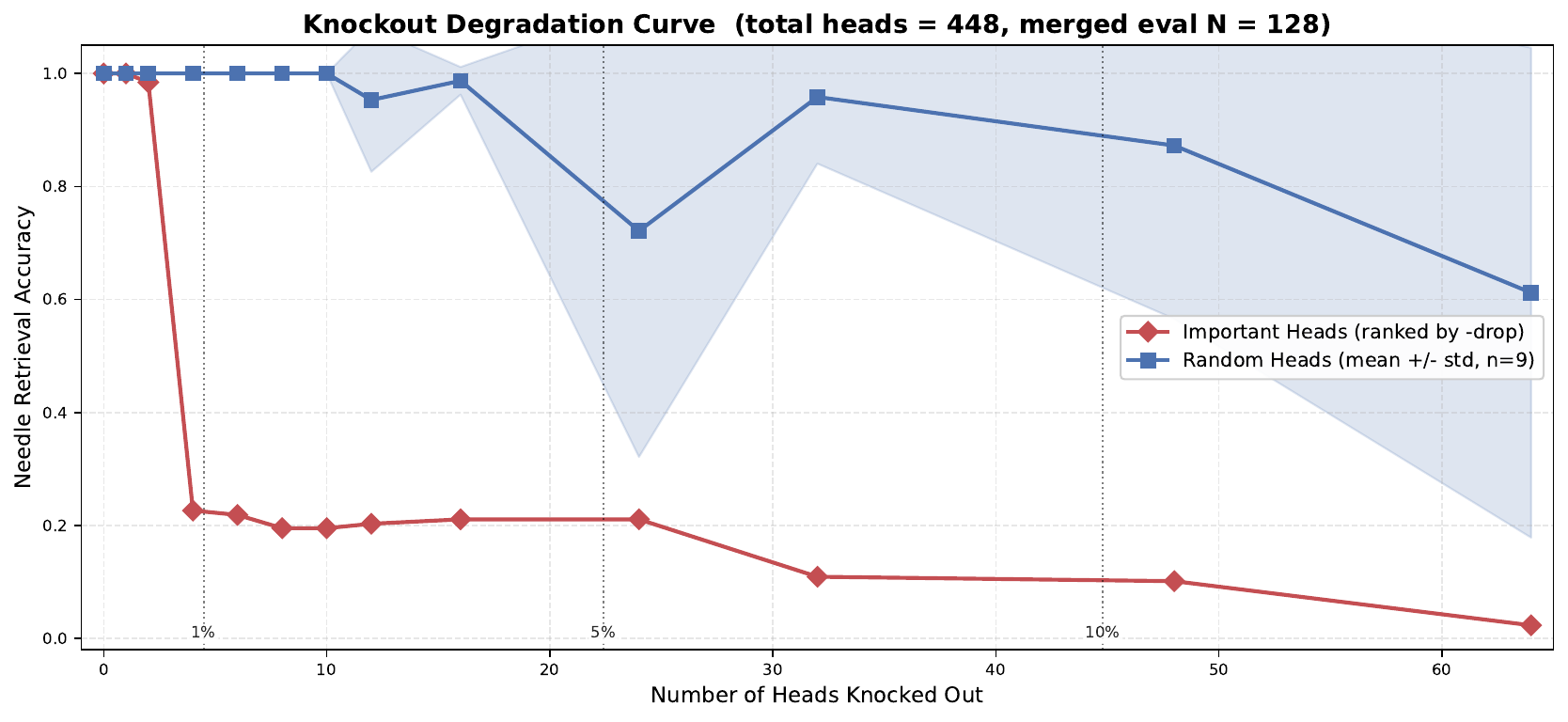}
        \caption{}%
        \label{fig:interp_knockout}
    \end{subfigure}
    \caption{Head-level localization and its causal validation on Qwen3-1.7B ($448$ heads). \textbf{(a)} Head-specialization map: each cell is a head, colored by role, with $+/-$ markers where importance increases/decreases with context length. Retrieval importance is localized to scattered \emph{heads}, not whole layers. \textbf{(b)} Knocking faithfulness: ablating heads by estimated importance ($-\mathrm{drop}$) collapses needle-retrieval accuracy after only the top few heads ($\approx1\%$), whereas random controls (mean $\pm$ std) stay far higher at every budget.}%
    \label{fig:interp_results}
\end{figure}

\paragraph{Retrieval is head-localized, not layer-localized.}
This is the central observation behind head-level hybridization. If long-context retrieval were a \emph{layer}-level property---the implicit assumption of layer-wise hybrids---then within any critical layer all $16$ heads would carry comparable importance, and a clean classification of heads into long-context vs. replaceable would make each layer \emph{pure}. Neither holds. The per-layer Gini coefficient of head importance averages $0.622$ (ranging $0.399$ to $0.915$): importance is concentrated in a few heads even within the most critical layers. Classifying heads by their average importance over context lengths yields only $29$ long-context-critical heads ($\lvert\mathrm{drop}\rvert\geq0.01$) against $407$ replaceable ones ($<0.005$)---i.e., only $\approx6.5\%$ of the $448$ heads are essential for retrieval, while $\approx90.8\%$ are safely convertible to GDN. Crucially, these critical heads are \emph{scattered}: the head-specialization map (Figure~\ref{fig:interp_granularity}) shows that almost every layer is a \emph{mixture} of critical and replaceable heads rather than a single role, with ten layers each containing both a long-context-critical head and several replaceable ones (enumerated in Table~\ref{tab:counterexample_layers}). Any per-layer assignment must therefore either waste FA on unimportant heads or convert away critical ones; only head-level allocation can separate them. This causal, importance-based argument---not representational fineness---is why the head is the right axis for hybridization.

\paragraph{Causal faithfulness under knockout.}
The importance score is a \emph{soft} logit-difference measurement; we validate it against the \emph{hard} downstream behavior. We knock out heads in decreasing order of estimated importance (ranked by $-\mathrm{drop}$) by replacing their output with the corrupted-run activation during prefill, then greedily generate and score needle-retrieval accuracy; as controls we knock out the same number of randomly chosen heads over several random runs. As shown in Figure~\ref{fig:interp_knockout}, removing only the top few importance-ranked heads collapses retrieval accuracy from near-perfect to a small fraction, and it falls further toward zero as more are removed, whereas the random controls remain far higher at every budget and near-perfect over the first several knockouts. This confirms that the score identifies the heads that causally carry retrieval, not an arbitrary subset.

Together, these analyses explain the gains in Tables~\ref{tab:head_selection} and~\ref{tab:comp_ratio}: retrieval lives in a sparse, scattered, causally verifiable set of heads that no per-layer assignment can isolate, and interpretability-guided screening is precisely the tool that recovers it.

\subsection{Training configuration optimization}

\label{subsec:optim_training}

To maximize the potential of our head-wise hybrid architecture, we conducted a systematic optimization of the training configurations across the three-stage transfer learning pipeline. 

\paragraph{Experimental Setup.} While maintaining FineWeb-Edu as the core pre-training corpus, we focused on refining the data scale and hyperparameter schedules to enhance both convergence stability and final model performance. We conduct experiments based on the Default Model as denoted before.

\begin{table}[]
    \centering
    \caption{Optimized training configuration of three training stages.}%
    \label{tab:new_train_config}
    \begin{tabular}{c|ccccc}
        \toprule
        Stage &  Batch Size & Context Length & Training Steps (Tokens) & LR scheduler & Learning Rate \\
        \midrule
        1 & 128 & 2048 & 3052 (0.8B) & Cosine & 1e-3 $\rightarrow$ 1e-5\\
        2 & 256 & 2048 & 7630 (4.0B) & Cosine & 1e-4 $\rightarrow$ 1e-5\\
        3 & 128 & 16384 & 477 (1.0B) & Constant & 1e-5 \\
        \bottomrule
    \end{tabular}

\end{table}

\begin{table}[t]
    \centering
    \caption{Comparison of training configuration optimization. $^\dagger$ denotes that the results are reproduced by ourselves. $\text{New-Config}$ indicates the optimized configuration, which is summarized in Table~\ref{tab:new_train_config}.}%
    \label{tab:pip_config}
    \begin{tabular}{lcccccc}
        \toprule
        \multirow{2}{*}{Model} & \multicolumn{2}{c}{RULER Single} & \multicolumn{2}{c}{RULER Multi-Key} & \multicolumn{2}{c}{General Reasoning} \\
        \cmidrule(lr){2-3} \cmidrule(lr){4-5} \cmidrule(lr){6-7}
         & (Native) & (Extended) & (Native) & (Extended) & (Hard) & (Easy) \\
        \midrule
        HypeNet$^\dagger$  & \textbf{89.07} & \textbf{85.00} & 24.85 & 24.37 & 19.80 & 59.72 \\
        + New-Config & 86.57 & 76.62 & \textbf{33.85} & \textbf{25.30} & \textbf{23.71} & \textbf{60.87}\\
        \midrule
        Default Model & 98.47 & 87.49 & 37.10 & 27.37 & 31.03 & 62.12\\
        + New-Config & \textbf{99.40} & \textbf{89.84} & \textbf{51.00} & \textbf{35.13} & \textbf{34.91} & \textbf{63.14}\\
        \bottomrule
    \end{tabular}

\end{table}

\paragraph{Results and Analysis.} We optimize the training of the hybrid model transfer learning from multiple dimensions, including data scale and training configuration. The optimized configuration of three training stages is summarized in Table~\ref{tab:new_train_config}. Compared to the original setup, we observe that scaling up the training data for Stage 1 and Stage 2 significantly enhances the efficacy of knowledge transfer. Furthermore, we increase both the training batch size and the sequence length to better accommodate long-context dependencies and stabilize the optimization dynamics across the heterogeneous attention branches. Table~\ref{tab:pip_config} presents a detailed comparison of model performance before and after the training configuration optimization. The results highlight two key observations:

\textbf{Significant Performance Gains for Our Hybrid Model.}
Under the optimized configuration, our proposed head-wise hybrid architecture demonstrates consistent improvements across all six evaluation benchmarks. Notably, the performance gains are most pronounced in challenging scenarios, such as RULER Multi-key retrieval and complex general reasoning tasks. For instance, accuracy on RULER Multi-key increases by 13.90\% at the native context length, indicating a substantial improvement in handling intricate long-context dependencies. This robust improvement underscores the efficacy of our head-wise selection strategy in capturing fine-grained semantic interactions, validating that scaling up data in the alignment and distillation stages effectively empowers the model to internalize heterogeneous attention patterns.

\textbf{Divergent scaling behavior of HypeNet.}
In contrast, HypeNet exhibits marginal improvements under the optimized configuration. While it achieves moderate gains exceeding 4.00\% on Multi-Key retrieval within the native context length and on challenging general reasoning benchmarks, these improvements are overshadowed by a pronounced degradation on Single-key retrieval tasks, where performance drops by 2.50\% to 8.38\%. This asymmetry suggests that head-wise hybridization has a higher capacity for absorbing additional training data: the finer-grained FA/GDN assignment allows the model to better exploit the representational complementarity between branches, whereas HypeNet's layer-wise mixing offers less room for the model to redistribute learned representations across attention types.

\subsection{Scaling Up with More Training Tokens}

\label{subsec:scaling_up}

We further probe the potential of our hybrid models by scaling up the carefully filtered high-quality training data to over 15B tokens and benchmarking against a range of open-source models around 2B parameters—including standard Transformers and other hybrid architectures.

\paragraph{Experimental Setup.} For fair and rigorous comparison, all baseline models utilized in our evaluation are the official releases provided by their respective authors. We conduct all experiments within a unified evaluation framework, employing identical preprocessing pipelines and hyperparameter settings to eliminate implementation biases and guarantee the reproducibility of our results. To assess long-context retrieval capabilities across varying sequence lengths, we evaluate models on the NIAH benchmark, reporting the averaged results of both single-key and multi-key retrieval tasks over context lengths spanning from 4K to 256K tokens. For general reasoning, we deliberately select the hard subset to better expose performance differences among models. We benchmark against two categories of state-of-the-art architectures: (1) Standard Transformer-based models, including Qwen3-1.7B and MiniCPM-1B-Base~\cite{hu2024minicpmunveilingpotentialsmall}; and (2) Hybrid Architecture models, such as Gemma-3n-E2B~\cite{gemmateam2025gemma3technicalreport}, HypeNet, Hymba-1.5B~\cite{dong2024hymbahybridheadarchitecturesmall}, and Jet-Nemotron-2B~\cite{gu2025jet}.

\begin{table}[t]
    \centering
    \caption{Comparison with SOTA models on long-context retrieval benchmarks across context lengths ranging from 4K to 256K.}%
    \label{tab:sota_comparison_niah}
    \begin{tabular}{lcccccccc}
        \toprule
        Model & Type & 4K & 8K & 16K & 32K & 64K & 128K & 256K \\
        \midrule
        \multicolumn{9}{c}{RULER Single} \\
        \midrule
        Qwen3-1.7B & FA & \textbf{100.00} & 99.87 & 99.93 & \textbf{99.93} & 98.13 & 0.00 & 0.00\\
        Qwen3-1.7B-yarn & FA & \textbf{100.00} & \textbf{99.93} & 98.93 & 97.73 & 98.80 & 83.47 & 40.20 \\
        MiniCPM5-1B-Base & FA & 99.93 & 99.47 & 98.60 & 97.60 & 90.93 & 93.73 & 59.53\\
        Gemma-3n-E2B & Hybrid & 99.93 & 99.87 & 99.67 & 99.67 & 22.23 & 0.00 & 0.00\\
        HypeNet-2B & Hybrid & 93.60 & 94.53 & 86.33 & 84.33 & 83.93 & 78.80 & 68.93\\
        Hymba-1.5B & Hybrid & 95.60 & 88.20 & 25.60 & 0.07 & 0.00 & 0.00 & 0.00\\
        Jet-Nemotron-2B & Hybrid & \textbf{100.00} & 99.67 & 99.47 & 92.80 & 27.27 & 12.60 & 1.07\\
        \rowcolor{lightgray!30} HydraHead & Hybrid & \textbf{100.00} & \textbf{99.93} & \textbf{100.00} & 99.80 & \textbf{99.80} & \textbf{99.47} & \textbf{94.53}\\
        \midrule
        \multicolumn{9}{c}{RULER Multi-Key} \\
        \midrule
        Qwen3-1.7B & FA & \textbf{99.50} & \textbf{99.40} & \textbf{97.80} & \textbf{94.10} & 42.20 & 0.00 & 0.00\\
        Qwen3-1.7B-yarn & FA & 97.40 & 91.80 & 54.30 & 46.10 & 47.80 & 41.50 & 14.20\\
        MiniCPM5-1B-Base & FA & 98.10 & 96.60 & 89.90 & 87.20 & 76.20 & 54.40 & 16.60\\
        Gemma-3n-E2B & Hybrid & 98.70 & 98.50 & 89.70 & 76.30 & 11.70 & 0.00 & 0.00\\
        HypeNet-2B & Hybrid & 30.50 & 28.90 & 25.00 & 13.90 & 27.40 & 16.00 & 16.90\\
        Hymba-1.5B & Hybrid & 44.10 & 33.40 & 12.50 & 0.00 & 0.00 & 0.00 & 0.00\\
        Jet-Nemotron-2B & Hybrid & 95.70 & 89.90 & 69.00 & 42.00 & 8.80 & 1.90 & 0.00\\
        
        \rowcolor{lightgray!30} HydraHead & Hybrid & 98.50 & 96.40 & 94.30 & 92.50 & \textbf{82.00} & \textbf{68.80} & \textbf{52.70}\\
        \bottomrule
    \end{tabular}

\end{table}

\begin{table}[]
    \centering
    \caption{Comparison with SOTA models on the general benchmark (\textbf{Hard}).}%
    \label{tab:sota_comparison_hard}
    \begin{tabular}{lcccccc}
        \toprule
        Model & Type & MMLU & BBH & MBPP & GSM8k & Average\\
        \midrule
        Qwen3-1.7B & FA & 55.46 & 48.16 & 43.00 & 69.45 & 54.02\\
        Qwen3-1.7B-yarn & FA & 53.45 & 45.65 & 42.40 & 59.97 & 50.37\\
        MiniCPM5-1B-Base & FA & 39.30 & 37.81 & 33.20 & 34.12 & 36.11\\
        Gemma-3n-E2B & Hybrid & 53.82 & 42.47 & 35.00 & 25.85 & 39.29\\
        HypeNet-2B & Hybrid & 46.52 & 9.15 & 4.80 & 4.93 & 16.35\\
        Hymba-1.5B & Hybrid & 49.67 & 8.89 & 16.60 & 0.99\tablefootnote{Base model result; the original paper reports 58.76\% for the instruct-tuned variant. The score is lower than expected and may warrant further investigation.} & 19.04\\
        Jet-Nemotron-2B & Hybrid & 53.59 & \textbf{58.16} & \textbf{54.20} & \textbf{75.28} & \textbf{60.31}\\
        \rowcolor{lightgray!30} HydraHead & Hybrid & \textbf{55.80} & 46.64 & 47.20 & 52.84 & 50.62\\
        \bottomrule
    \end{tabular}
\end{table}

\paragraph{Results and Analysis.} We present quantitative results on long-context retrieval (Table~\ref{tab:sota_comparison_niah}) and general reasoning (Table~\ref{tab:sota_comparison_hard}), and summarize our findings in two aspects below.

\textbf{Superior Long-Context Extrapolation over Standard Transformers.} Compared to Qwen3-1.7B, HydraHead demonstrates strong extrapolation capability beyond 64K tokens. At 256K context length, it surpasses the YaRN-enhanced variant by +54.23\% on RULER Single and +38.50\% on RULER Multi-Key. Within the native 32K range, HydraHead remains competitive with the base model and even exceeds the YaRN version on certain tasks. On the general reasoning benchmark, HydraHead trails Qwen3-1.7B by only 3.40 percentage points on average—a marginal gap given that our model was adapted via merely 15B tokens of transfer training rather than full-scale pretraining. Moreover, it matches the YaRN-enhanced variant, indicating that the hybrid architecture itself does not compromise general-purpose capabilities.

\textbf{Robustness Advantage over Existing Hybrid Architectures.} While most hybrid models degrade to near-zero performance at 256K, HydraHead sustains 94.53\% (Single) and 52.70\% (Multi-Key). On the general benchmark, HydraHead outperforms all competing hybrid models by at least 10 percentage points on average, with one exception: Jet-Nemotron-2B achieves a higher average score (+9.69 points). However, this advantage comes at a steep cost to long-context ability---Jet-Nemotron-2B falls behind HydraHead by more than 50 percentage points at context lengths $\ge$64K. This stark contrast validates that our fine-grained head-wise allocation strategy not only preserves strong general reasoning abilities but also ensures superior scalability and reliability in long-context scenarios, offering a more balanced solution than current state-of-the-art hybrid models.

\section{Conclusion}

In this report, we present a head-wise hybrid attention architecture that converts a pre-trained Transformer into a hybrid model by organically integrating FA and LA at the head level. A head-wise scale-normalized mechanism is designed to effectively fuse the heterogeneous features from different attention mechanisms. Guided by interpretability analysis, we identify and allocate critical heads to the FA branch to preserve high-fidelity representations, while routing the remaining heads to the LA branch for efficient long-context extension. To enable effective transfer learning, we employ a three-stage training pipeline that combines parameter reuse initialization with distillation.

In controlled experiments under identical training configurations, our head-wise hybrid architecture achieves over 10\% improvement on challenging general reasoning benchmarks compared to layer-wise hybrid alternatives, while also consistently outperforming them across long-context evaluations. Scaling the training to over 15B tokens, our model comprehensively surpasses the baseline, demonstrating strong potential for further scaling.

Several directions remain open for future work, including deeper exploration of interpretability-driven head allocation strategies, scaling studies across larger model and data sizes, and the integration of additional attention variants into the hybrid architecture. We hope these findings offer valuable insights to both academia and industry on the design and deployment of efficient hybrid attention models.

\bibliography{reference}
\bibliographystyle{unsrt}
\newpage

\appendix

\section{More Details of Various Hybrid Architectures}
\label{apd: hybrid_arch}

\begin{figure}
    \centering
    \includegraphics[width=1\linewidth]{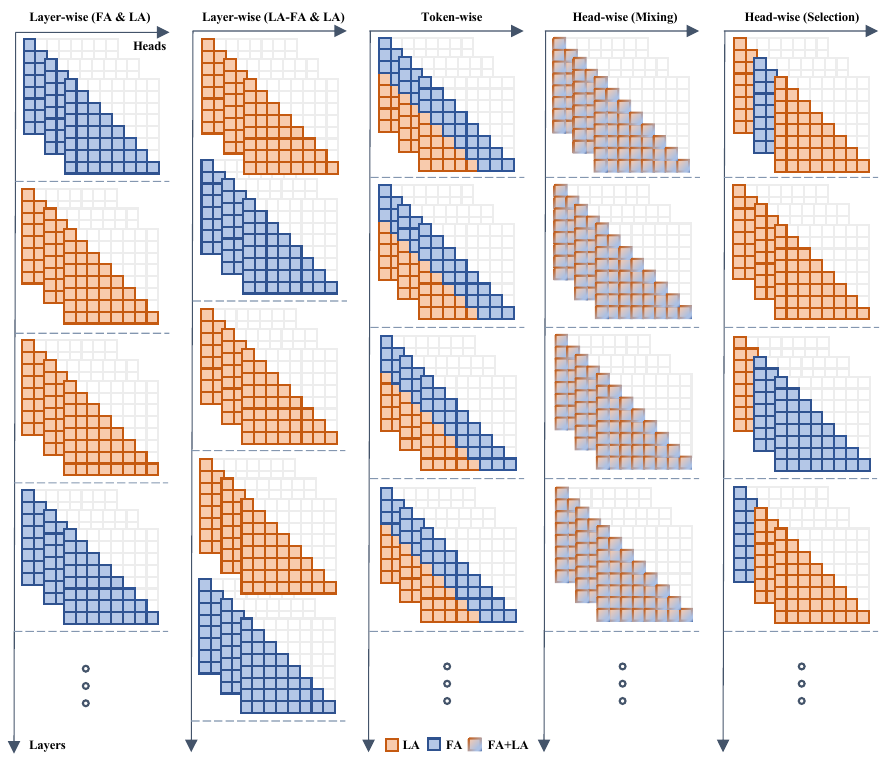}
    \caption{Visualization of some typical hybrid methods. The hybrids can be categorized in terms of layer-wise, token-wise, and head-wise.}%
    \label{fig:bybrid_cache_comp}
\end{figure}

To comprehensively evaluate the efficacy of different hybridization strategies, we construct and compare three representative architectures: layer-wise hybrids, token-wise hybrids, and head-wise hybrids (Figure~\ref{fig:bybrid_cache_comp}). All models incorporate the structural optimizations proposed in our main paper, including the addition of a gating branch to Full Attention (FA) layers with scaling mechanisms replacing RoPE, and the integration of RoPE positional encoding into Gated DeltaNet (GDN) layers.

\paragraph{Layer-wise Hybrid Architectures.}
We explore two distinct granularities for layer-wise mixing, guided by the layer selection masks derived from HypeNet. (a) \textbf{FA \& LA:} Each transformer layer is configured as either an FA block (FA + MLP) or a Linear block (GDN + MLP). The HypeNet-derived mask determines which layers retain FA capabilities. (b) \textbf{LA-FA \& LA:} Each layer is configured as either a Composite block (GDN + FA + MLP) or a Linear block (GDN + MLP). In this setting, the HypeNet mask identifies layers that require the combined power of both attention mechanisms, while others rely solely on GDN for efficiency.

\paragraph{Token-wise Hybrid Architectures.}
As illustrated in Figure~\ref{fig:bybrid_cache_comp}, our token-wise design conceptually assigns FA (implemented as Sliding Window Attention, SWA) to tokens within the recent window and LA to those outside it. For implementation simplicity, we realize this by processing all tokens through the GDN branch for global context modeling while additionally attending to recent tokens within the sliding window via FA to capture high-precision local dependencies.
We consider two variants of this design. The first applies 4K-context SWA combined with GDN uniformly across all layers. The second adopts a layer-interleaved strategy inspired by Liger~\cite{lan2025liger}: 75\% of the layers use SWA (window size 128) with GDN, while the remaining 25\% combine full FA with GDN. Unlike Liger, which employs pure FA in its global layers, our variant retains GDN alongside FA to preserve token-wise hybridization throughout all layers.

Although this design can be categorized as token-wise mixing due to the varying receptive fields across token positions, the way features from the two branches are combined is essentially identical to the head-wise mixing approach. A more canonical form of token-wise mixing operates at the KV cache level during softmax attention computation, where different tokens attend to KV caches of varying lengths or compositions~\cite{meng2026stillselectingtokensintralayer,zhang2025lolcatslowranklinearizinglarge}. However, our preliminary experiments reveal that such methods struggle to acquire long-context capabilities through transfer learning when the pretraining context window is orders of magnitude shorter than the target evaluation length. We therefore defer their comparison to future work and do not include them in this paper.

\paragraph{Head-wise Hybrid Architectures.} We define two sub-paradigms for head-wise hybridization: \textbf{head-wise mixing}, where each head simultaneously computes both FA and GDN branches and fuses their outputs, and \textbf{head-wise selection}, where each head is exclusively assigned to either the FA or GDN branch based on a learned or heuristic criterion.
For head-wise mixing, we implement a variant inspired by~\cite{dong2024hymbahybridheadarchitecturesmall} that processes the input through parallel FA and GDN branches, each operating over the full set of heads, and fuses their outputs at the feature level within every layer.
For head-wise selection, existing work has explored per-head hybridization between FA and sparse attention~\cite{xiao2024duoattentionefficientlongcontextllm,tang2026elasticattentiontesttimeadaptive}. We therefore use HydraHead as the representative of this paradigm that adopts linear attention instead.

\section{More Results of Hybrid Architecture Comparison}

We perform a detailed comparison of various typical hybrid architectures in Table~\ref{tab:arc_comparison}. Table~\ref{tab:arc_ruler} summarizes the long-context evaluation results on the RULER benchmark, covering both Single and Multi-key scenarios across context lengths ranging from 16K to 256K tokens. Our method achieves the best performance in nearly all settings, demonstrating robust and broad superiority in long-context modeling.
Tables~\ref{tab:arc_general_easy} and~\ref{tab:arc_general_hard} present the general capability comparisons. Our approach significantly surpasses layer-wise hybrid models, with particularly notable gains in mathematical reasoning and code generation. While other head-wise and token-wise hybrid variants achieve stronger general performance than our model, they suffer from severely compromised long-context capabilities. Taken together, our hybrid design achieves the most favorable trade-off between general competence and long-context proficiency.

\begin{table}[]
    \centering
    \caption{Comparison of various hybrid architectures in the \textbf{RULER} benchmark. $\text{Cache}$ denotes the relative key-value cache under a 16K context length, where the standard FA cache is normalized to 1.0. 
    $\text{Light.}$ denotes the lightning attention~\cite{qin2024variouslengthsconstantspeed}. 
    $^*$ and $^\ddagger$ indicate the hybrids are similar to HypeNet~\cite{chen2026hybridlinearattentionright} and Liger~\cite{lan2025liger}, respectively. $^\dagger$ indicates that an FA layer is inserted after every 3 LA layers. HW denotes Head-wise.}%
    \label{tab:arc_ruler}
    \begin{tabular}{lccccccccccc}
        \toprule
        \multirow{2}{*}{Model} & \multirow{2}{*}{LA} & \multicolumn{5}{c}{RULER Single} & \multicolumn{5}{c}{RULER Multi-Key}\\
        \cmidrule(lr){3-7} \cmidrule(lr){8-12}
         & & 16K & 32K & 64K & 128K & 256K & 16K & 32K & 64K & 128K & 256K \\
        \midrule
        \multicolumn{12}{c}{Layer-wise Hybrid Transformers} \\
        \midrule
        FA \& LA$^*$ & Light. & 92.13 & 86.00 & 88.87 & 84.47 & \textbf{81.67} & 28.90 & 20.80 & \textbf{32.70} & 22.70 & 17.70\\
        FA \& LA$^\dagger$ & Light. & 87.67 & 84.40 & 80.33 & 69.40 & 53.53 & 36.70 & 26.70 & 26.00 & 14.60 & 12.80\\
        FA \& LA & GDN & 62.87 & 32.40 & 15.67 & 4.20 & 1.20 & 25.10 & 19.90 & 10.30 & 4.80 & 1.50\\
        LA-FA \& LA & GDN & 73.13 & 43.87 & 17.27 & 5.20 & 1.93 & 23.00 & 15.00 & 12.80 & 5.70 & 2.00\\
        \midrule
        \multicolumn{12}{c}{Token-wise \& Head-wise Hybrid Transformers} \\
        \midrule
        Token-wise & GDN & 28.07 & 13.47 & 6.07 & 3.00 & 2.13 & 21.60 & 10.50 & 3.40 & 2.40 & 1.50\\
        Token-wise$^\ddagger$ & GDN & 97.87 & 94.20 & 80.87 & 56.13 & 39.33 & 29.00 & 23.40 & 15.70 & 9.60 & 7.80\\
        HW Mixing & GDN & 98.33 & 88.07 & 77.73 & 61.73 & 41.80 & 46.10 & 28.60 & 21.30 & 14.30 & 8.00\\
        HydraHead & GDN & \textbf{98.60} & \textbf{98.33} & \textbf{94.60} & \textbf{89.00} & 78.87 & \textbf{41.40} & \textbf{32.80} & 32.30 & \textbf{26.60} & \textbf{23.20}\\
        \bottomrule
    \end{tabular}

\end{table}

\begin{table}[]
    \centering
    \caption{Comparison of various hybrid architectures in the general benchmark (\textbf{Easy}). $\text{Cache}$ denotes the relative key-value cache under a 16K context length, where the standard FA cache is normalized to 1.0. 
    $\text{Light.}$ denotes the lightning attention~\cite{qin2024variouslengthsconstantspeed}. 
    $^*$ and $^\ddagger$ indicate the hybrids are similar to HypeNet~\cite{chen2026hybridlinearattentionright} and Liger~\cite{lan2025liger}, respectively. $^\dagger$ indicates that an FA layer is inserted after every 3 LA layers.}%
    \label{tab:arc_general_easy}
    \begin{tabular}{lccccccc}
        \toprule
        Model & LA & ARC-C & ARC-E & HellaSwag & LAMBADA & PIQA & Winogrande \\
        \midrule
        \multicolumn{8}{c}{Layer-wise Hybrid Transformers} \\
        \midrule
        FA \& LA$^*$ & Light. & 41.38 & 71.13 & 58.81 & 52.88 & 72.74 & 61.40\\
        FA \& LA$^\dagger$ & Light. & 43.52 & 72.01 & 59.20 & 50.86 & 72.85 & 63.30\\
        FA \& LA & GDN & 39.85 & 70.75 & 58.05 & 47.25 & 71.76 & 61.40\\
        LA-FA \& LA & GDN & 41.21 & 70.24 & 58.46 & 49.93 & 72.03 & 60.30\\
        \midrule
        \multicolumn{8}{c}{Token-wise \& Head-wise Hybrid Transformers} \\
        \midrule
        Token-wise & GDN & \textbf{46.50} & 75.38 & \textbf{62.58} & \textbf{58.49} & 74.05 & 63.38\\
        Token-wise$^\ddagger$ & GDN & 45.65 & 75.34 & 62.43 & 58.28 & \textbf{74.27} & \textbf{64.25}\\
        Head-wise Mixing & GDN & 43.60 & \textbf{76.09} & 61.97 & 57.33 & 73.61 & 64.01\\
        HydraHead & GDN & 42.83 & 76.05 & 61.82 & 56.32 & 73.88 & 61.80\\
        \bottomrule
    \end{tabular}

\end{table}

\begin{table}[]
    \centering
    \caption{Comparison of various hybrid architectures in the general benchmark (\textbf{Hard}). $\text{Cache}$ denotes the relative key-value cache under a 16K context length, where the standard FA cache is normalized to 1.0. 
    $\text{Light.}$ denotes the lightning attention~\cite{qin2024variouslengthsconstantspeed}. 
    $^*$ and $^\ddagger$ indicate the hybrids are similar to HypeNet~\cite{chen2026hybridlinearattentionright} and Liger~\cite{lan2025liger}, respectively. $^\dagger$ indicates that an FA layer is inserted after every 3 LA layers.}%
    \label{tab:arc_general_hard}
    \begin{tabular}{lccccc}
        \toprule
        Model & LA & MMLU & BBH & MBPP & GSM8k\\
        \midrule
        \multicolumn{6}{c}{Layer-wise Hybrid Transformers} \\
        \midrule
        FA \& LA$^*$ & Light. & 48.80 & 13.79 & 10.00 & 6.60\\
        FA \& LA$^\dagger$ & Light. & 39.70 & 9.77 & 7.60 & 6.67\\
        FA \& LA & GDN & 46.67 & 13.58 & 2.40 & 3.11\\
        LA-FA \& LA & GDN & 48.32 & 14.53 & 4.40 & 6.44\\
        \midrule
        \multicolumn{6}{c}{Token-wise \& Head-wise Hybrid Transformers} \\
        \midrule
        Token-wise & GDN & \textbf{56.65} & \textbf{47.87} & \textbf{34.60} & \textbf{50.11}\\
        Token-wise$^\ddagger$ & GDN & 53.17 & 27.65 & 29.20 & 36.63\\
        Head-wise Mixing & GDN & 54.46 & 35.94 & 29.20 & 32.68\\
        Ours & GDN & 49.86 & 29.15 & 22.00 & 23.12\\
        \bottomrule
    \end{tabular}

\end{table}

\section{Head-Selection Details and Additional Interpretability Analysis}
\label{apd:interp}

This appendix collects the implementation details of the head-importance estimator of Section~\ref{sec:head_selection}, the robustness caveats referenced there, the second-capability (general-ability) localization, and the full set of interpretability figures for Qwen3-1.7B summarized in Section~\ref{subsec:interp_anatomy}.

\subsection{Estimator details and caveats}
\label{apd:interp_details}

\paragraph{Direct-effect measurement.}
When scoring receivers (Equation~\eqref{eq:ie}) we freeze every \emph{downstream} attention output to its clean value, so the patched head can influence the logits only through the residual stream and the downstream MLPs. This isolates the head's \emph{direct effect} on the readout, in the sense of Wang et al.~\cite{wang2022interpretabilitywildcircuitindirect}, rather than a total effect that would also route through downstream attention. The downstream MLPs are intentionally left unfrozen, so the measurement captures the head's contribution as actually consumed by the rest of the network.

\paragraph{Corruption design.}
For the retrieval capability the counterfactual $x'$ is a \emph{symmetric token replacement}: the needle value is replaced by a fresh distractor of the same type and the same token length, with the haystack, query, and structure held identical. This keeps $x'$ on the model's natural distribution---the recommended practice over additive activation noise, which pushes the residual stream out of distribution and propagates the artifact downstream~\cite{zhang2024towardsbestpractices}. The span decay in Equation~\eqref{eq:logit_diff} is $\lambda=0.9$ for the retrieval probe.

\paragraph{Calibration configuration.}
We localize on a small calibration set---eight NIAH samples in our experiments---drawn at short context, and transfer the resulting selection to long context, which is justified by the length-robustness of the selected set (Figure~\ref{fig:interp_validity}). The per-head ranking is already stable from roughly six samples (Figure~\ref{fig:interp_validity}), so the exact sample count is not load-bearing. Scores are computed in \texttt{fp16}: this preserves the importance ranking, whereas lower-precision \texttt{bf16} can collapse genuinely distinct heads to indistinguishable scores. We aggregate importance as the union of the per-head scores over the five NIAH sub-probes and mark a head as critical when $\lvert\mathrm{drop}\rvert \ge 0.01$.

\paragraph{Patching engine.}
We use full activation patching, implemented in two phases for tractability: a layer-level screen (patch all heads of a layer at once) followed by a head-level refinement of the surviving layers. No gradients or attribution approximations are used in the reported numbers. The scalable alternative, attribution patching, estimates all heads' effects from a single backward pass via a first-order Taylor expansion of Equation~\eqref{eq:ie}; its linearization is accurate enough for the rank-ordered top-$K$ use we make of it (its \emph{exploratory} regime), though not for absolute effect sizes~\cite{nanda2023attributionpatching}.

\paragraph{Backup heads and rank noise.}
Single-head necessity \emph{under}-estimates heads that have redundant ``backup'' pathways: knocking one out lets a backup compensate, so the measured drop is smaller than the head's true role~\cite{wang2022interpretabilitywildcircuitindirect}. Individual head ranks are therefore noisy even when the selected \emph{set} is robust, which is why we (i) take the union across sub-probes and (ii) rely on the head population rather than exact ranks (Section~\ref{subsec:interp_anatomy}). The selection step is exploratory (rank order); the knockout study (Figure~\ref{fig:interp_knockout}) is the confirmatory check that the chosen set is causally load-bearing.

\subsection{Functional diversity of heads within a layer}
\label{apd:interp_diversity}

\begin{figure}[t]
    \centering
    \includegraphics[width=0.56\linewidth]{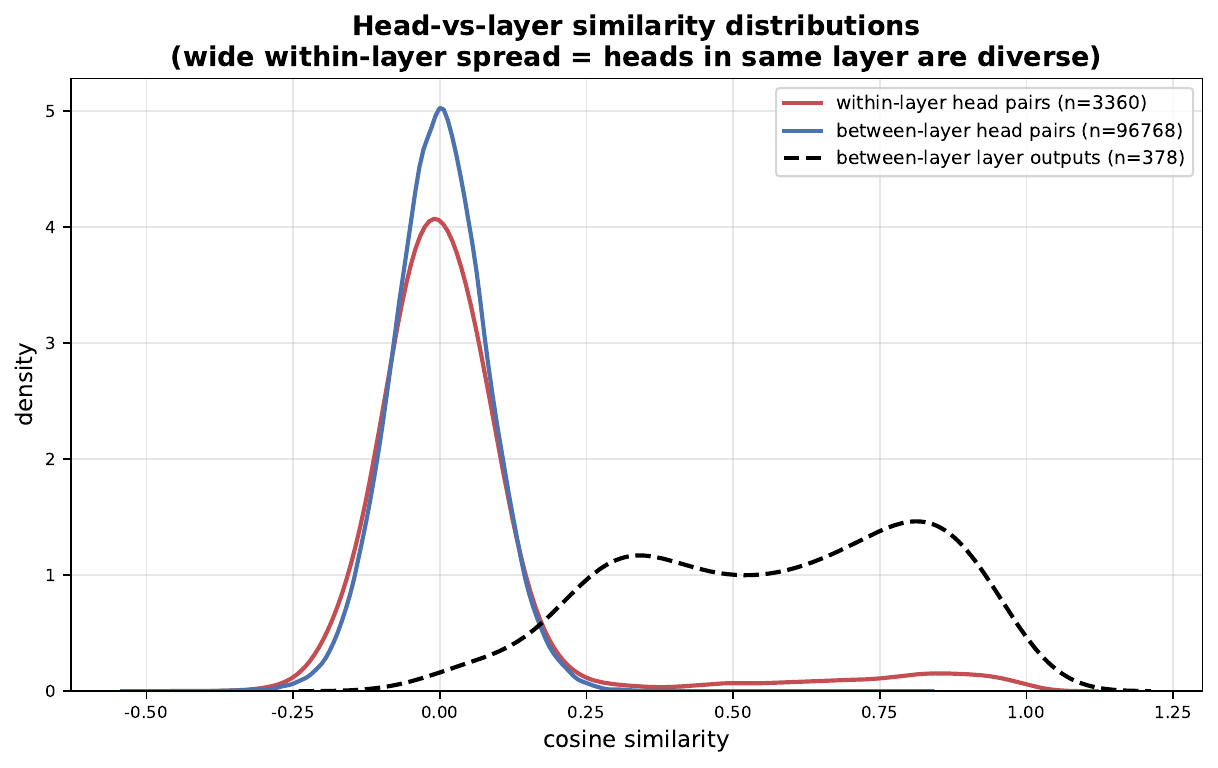}\hfill
    \includegraphics[width=0.42\linewidth]{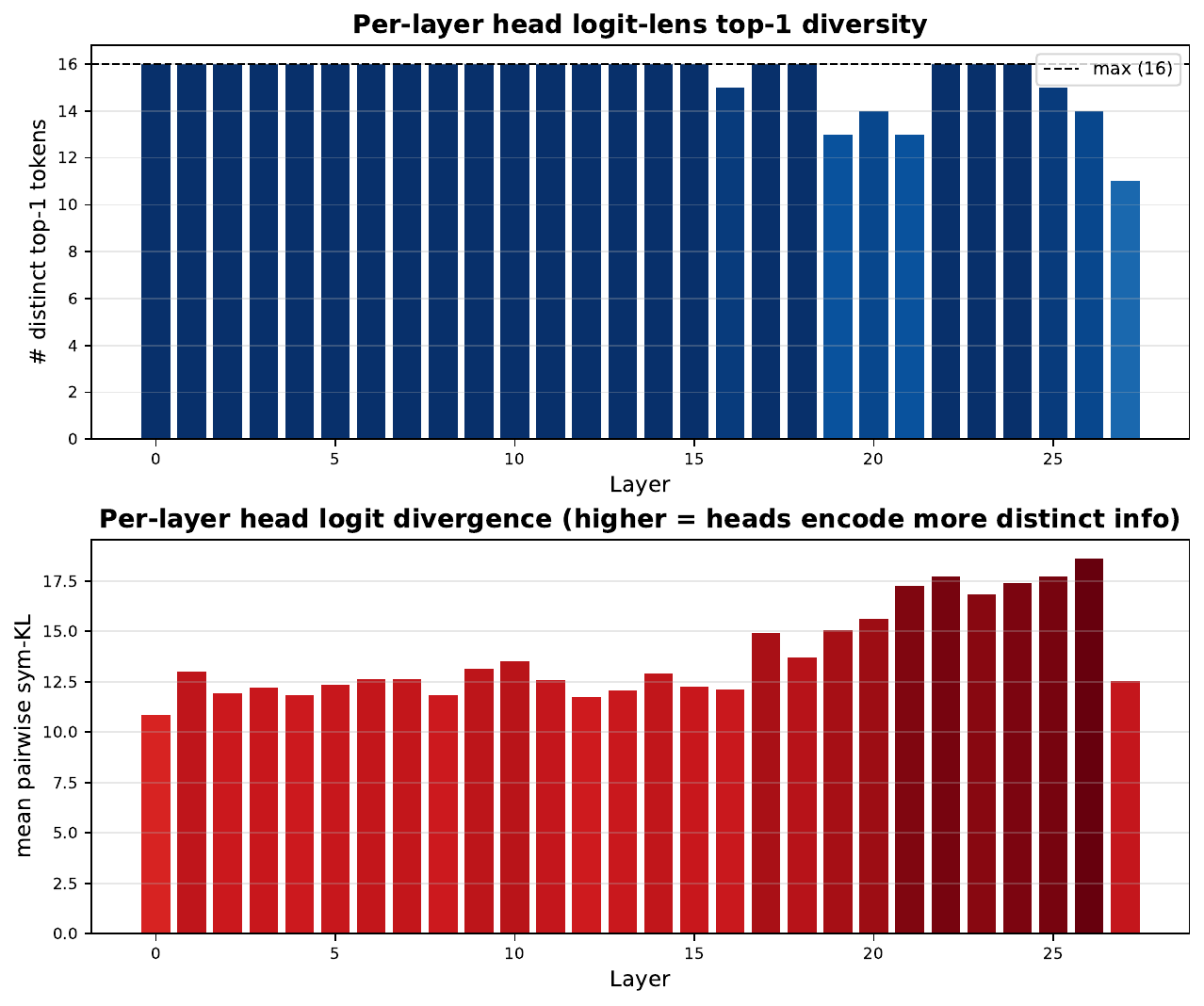}
    \caption{Functional diversity of heads within a layer (Qwen3-1.7B). \textbf{(Left)} Distributions of cosine similarity for within-layer head pairs vs. between-layer head pairs vs. between-layer full-output pairs; within-layer head pairs span a wide range, indicating non-redundancy. \textbf{(Right)} Per-layer count of distinct top-$1$ logit-lens tokens across the $16$ heads (max $16$) and their mean pairwise symmetric KL; both are high in mid/late layers.}%
    \label{fig:interp_repr}
\end{figure}

\begin{figure}[t]
    \centering
    \includegraphics[width=0.62\linewidth]{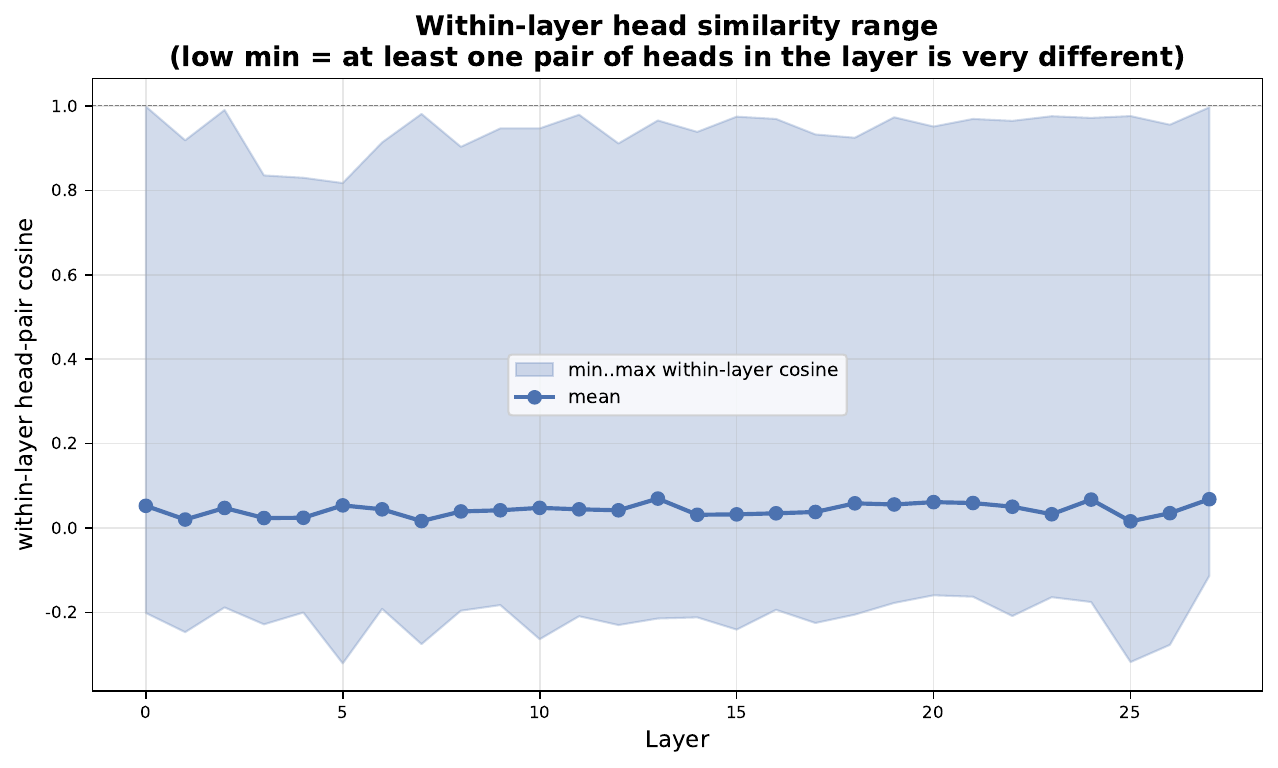}
    \caption{Per-layer (min, mean, max) within-layer head-pair cosine similarity on Qwen3-1.7B; the minimum dips to near zero for many layers, i.e.\ at least one pair of heads in such a layer encodes near-orthogonal information. The full layer-level ($28\times28$) similarity matrix is shown in Figure~\ref{fig:comm_layersim}.}%
    \label{fig:apx_repr}
\end{figure}

The head-localization result of Section~\ref{subsec:interp_anatomy} presumes that heads in a layer actually do different things, which we verify directly. Comparing the $448\times448$ matrix of per-head output cosine similarities against the $28\times28$ matrix of full layer-output similarities (Figure~\ref{fig:apx_repr} and Figure~\ref{fig:comm_layersim}), the within-layer head-pair similarities span a wide range---down to near-orthogonal for at least one pair in most layers (Figure~\ref{fig:apx_repr})---whereas the layer-level matrix is smooth. A head-level logit lens tells the same story: projecting each head's contribution to vocabulary space, the $16$ heads of a mid-to-late layer predict many \emph{distinct} top-$1$ tokens and have high pairwise symmetric KL divergence (Figure~\ref{fig:interp_repr}). Heads inside a single layer are therefore not redundant copies, and the per-head granularity carries information invisible at the layer level---the representational counterpart of the causal head-localization result and the empirical claim underlying Figure~\ref{fig:comm_visual}.

\begin{table}[t]
    \centering
    \caption{Counter-examples to layer-level granularity on Qwen3-1.7B: the top layers that simultaneously contain long-context-critical heads ($\lvert\mathrm{drop}\rvert\geq0.01$) and replaceable heads ($<0.005$). Every such layer would be mis-served by a uniform per-layer mechanism assignment.}%
    \label{tab:counterexample_layers}
    \begin{tabular}{cccccccccc}
        \toprule
        Layer & 27 & 23 & 20 & 21 & 25 & 26 & 19 & 22 & 24 \\
        \midrule
        \# long-context heads & 6 & 4 & 3 & 3 & 3 & 3 & 2 & 2 & 2 \\
        \# replaceable heads  & 9 & 12 & 11 & 10 & 13 & 12 & 13 & 14 & 13 \\
        \bottomrule
    \end{tabular}
\end{table}

\section{Limitations and Future Work}

While HydraHead demonstrates promising results in balancing general reasoning and long-context retrieval, several limitations point to directions for future exploration.

\textbf{Model scale and architecture diversity.} Our experiments are primarily conducted on Qwen3-1.7B, a relatively compact dense Transformer. Despite this modest scale, our results provide compelling evidence for the potential of head-wise hybridization as a general design principle—the consistent gains observed across both retrieval-centric and reasoning benchmarks suggest that functional specialization of attention heads is a robust phenomenon worth exploiting at larger scales. Validating this principle on larger-scale models (e.g., 7B, 13B, and beyond) and across diverse architectures—including Mixture-of-Experts and multimodal models—remains an important next step. The functional specialization patterns of attention heads may shift with model scale and modality, and understanding how these shifts interact with our interpretability-guided head selection strategy would further strengthen the generality of the approach.

\textbf{Training data scale and post-training.} The current scaling experiments are limited to approximately 15B tokens, which is modest by contemporary standards. Scaling up the training corpus by an order of magnitude or more, combined with post-training stages such as instruction tuning and reinforcement learning from human feedback, could unlock substantially stronger hybrid models. In particular, it remains to be seen whether head-wise hybridization confers additional benefits in alignment-centric post-training regimes, where the balance between precise instruction following and long-context reasoning becomes especially critical.

\textbf{Scalability and generality of interpretability-guided selection.} The current head selection procedure relies on full activation patching, which requires iterating over all attention heads with separate forward passes per head---tractable at the 1.7B scale but increasingly costly at frontier scales. Nevertheless, even full activation patching remains far cheaper than the alternative of training multiple hybrid models under different head configurations to search for a good assignment. While attribution patching offers a single-backward-pass approximation, its accuracy for this specific use case has not been systematically validated in our experiments. Beyond computational cost, the procedure also depends on carefully designed calibration inputs: the paired counterfactual construction, the choice of readout metric, and the target capability specification (NIAH sub-probes as a proxy for general long-context retrieval) all require domain-informed design decisions that may not transfer directly to new capabilities or task families. That said, this design effort is a one-time cost amortized across all downstream hybrid models built from a given checkpoint, and the resulting selection is stable and transfers across context lengths. Developing more automated or task-agnostic selection protocols would further broaden the applicability of interpretability-guided hybridization.

\textbf{Gap between interpretability-based importance and deployable FA budget.} Our interpretability analysis suggests that only a small fraction of heads are highly important: roughly 6.5\% of heads are identified as carrying the most critical retrieval-related signal. However, the ratio experiments indicate that when the retained FA budget is reduced to around 10\%, model capability already deteriorates noticeably. This mismatch suggests that the current interpretability signal, while clearly informative, is not yet sufficient to fully determine the minimal FA allocation required for strong end-task performance. One possibility is that the head-importance estimation itself remains incomplete or noisy, and could be improved with better causal analysis or more robust calibration procedures. Another possibility is that the architecture and training pipeline are not yet able to preserve capability under such aggressive compression, even when the most important heads are retained. Addressing this gap may therefore require progress on both fronts: more accurate interpretability-guided selection, and stronger hybrid architectures or optimization strategies that can better exploit extremely sparse FA budgets.

\textbf{Broader attention mechanism combinations.} It is important to emphasize that the head-wise hybridization architecture we propose is not inherently tied to any specific attention mechanism—it is a general architectural paradigm that can accommodate arbitrary combinations of attention variants. In this work, we instantiate the architecture with two representative mechanisms, FA and GDN, as a concrete proof of concept. The architecture naturally extends to a much richer design space. On the efficient attention side, beyond LA variants (e.g., Mamba, RWKV, Linear Transformer), FA can also be combined with sparse attention mechanisms (e.g., sliding window, local attention) or latent-space attention formulations, each offering distinct efficiency–expressiveness trade-offs. More ambitiously, the architecture supports functional hybridization of more than two attention types—for instance, jointly mixing FA, LA, and sparse attention within a single model, where different heads adopt different mechanisms according to their functional roles. Systematically characterizing the head-level compatibility patterns across this rich design space could lead to more effective and deployment-specific hybrid recipes.

\end{document}